\documentclass{ieeeaccess}
\usepackage{cite}
\usepackage{amsmath,amssymb,amsfonts}
\usepackage{algorithmic}
\usepackage{graphicx}
\usepackage{textcomp}

\usepackage{bm}

\usepackage{cite}
\usepackage{booktabs}

\usepackage{url}
\usepackage{subfigure}

\usepackage{algorithm}

\makeatletter
\AtBeginDocument{\DeclareMathVersion{bold}
\SetSymbolFont{operators}{bold}{T1}{times}{b}{n}
\SetSymbolFont{NewLetters}{bold}{T1}{times}{b}{it}
\SetMathAlphabet{\mathrm}{bold}{T1}{times}{b}{n}
\SetMathAlphabet{\mathit}{bold}{T1}{times}{b}{it}
\SetMathAlphabet{\mathbf}{bold}{T1}{times}{b}{n}
\SetMathAlphabet{\mathtt}{bold}{OT1}{pcr}{b}{n}
\SetSymbolFont{symbols}{bold}{OMS}{cmsy}{b}{n}
\renewcommand\boldmath{\@nomath\boldmath\mathversion{bold}}}
\makeatother

\def\BibTeX{{\rm B\kern-.05em{\sc i\kern-.025em b}\kern-.08em
    T\kern-.1667em\lower.7ex\hbox{E}\kern-.125emX}}

\begin{document}

\bibliographystyle{IEEEtran}

\history{Date of publication xxxx 00, 0000, date of current version xxxx 00, 0000.}
\doi{}

\title{Enhancing Trust in Autonomous Agents: An Architecture for Accountability and Explainability through Blockchain and Large Language Models}
\author{\uppercase{Laura Fernández-Becerra},
\uppercase{Miguel Ángel González-Sanatamarta}, \uppercase{Ángel Manuel Guerrero-Higueras}, \IEEEmembership{Member, IEEE}, \uppercase{Francisco Javier Rodríguez-Lera}, \IEEEmembership{Member, IEEE}, \uppercase{Vicente Matellán-Olivera}, \IEEEmembership{Member, IEEE}}

\address{Robotics Group, University of León, León, 24071 S (e-mail: inflfb00@estudiantes.unileon.es)}

\tfootnote{This research has been partially funded by the Recovery, Transformation, and Resilience Plan, financed by the European Union (Next Generation) thanks to the TESCAC project (Traceability and Explainability in Autonomous Systems for improved Cybersecurity) granted by INCIBE to the University of León; and by grant PID2021-126592OB-C21 funded by MCIN/AEI/10.13039/501100011033 EDMAR (Explainable Decision Making in Autonomous Robots) project, PID2021-126592OB-C21 funded by MCIN/AEI/10.13039/501100011033 and by ERDF 'A way of making Europe'.}

\markboth
{L. Fernández-Becerra \headeretal: Preparation of Papers for IEEE TRANSACTIONS and JOURNALS}
{L. Fernández-Becerra \headeretal: Preparation of Papers for IEEE TRANSACTIONS and JOURNALS}

\corresp{Corresponding author: Laura Fernández Becerra (e-mail: inflfb00@estudiantes.unileon.es).}

\begin{abstract}
The deployment of autonomous agents in environments involving human interaction has increasingly raised security concerns. Consequently, understanding the circumstances behind an event becomes critical, requiring the development of capabilities to justify their behavior to non-expert users. Such explanations are essential to fostering trust and ensuring safety. Additionally, they contribute to improving communication by bridging the gap between the agent and the user, thereby enhancing the effectiveness of their interactions. This paper presents an accountability and explainability architecture implemented for mobile Robot Operating System-based robots. The proposed solution comprises two main components. First, a black box-like element used to provide accountability, featuring anti-tampering properties achieved through blockchain technology. Second, a component in charge of generating natural language explanations by harnessing the capabilities of Large Language Models over the data contained within the previously mentioned black box. This study evaluates the performance of our solution in three different scenarios, each involving autonomous agent navigation functionalities. This evaluation includes a thorough examination of accountability and explainability metrics, demonstrating the effectiveness of our approach in using accountable data from robot actions to obtain coherent, accurate and understandable explanations, even when facing the challenges inherent in the use of autonomous agents in real-world scenarios.
\end{abstract}


\titlepgskip=-21pt

\maketitle

\section{Introduction}
\label{sec:introduction}
The increasing integration of mobile robots into human environments has raised significant concerns regarding their transparency and trustworthiness. A key challenge in this context is ensuring that robots can provide meaningful justifications for their actions and decisions, particularly in safety-critical scenarios where their behavior may impact human well-being \cite{Sado2023}. Transparent and reliable explanations are essential \cite{borgo2018}, as they not only support effective human-robot interactions but also help build trust, particularly among non-expert users who depend on clear, accessible interpretations of robotic behavior \cite{Rosenfeld2019,Cruz2023}. However, the complexity of robotic decision-making processes and the need for thorough post-event analysis make it particularly difficult to achieve these goals.

To address these challenges, accountability must be ensured using a robust tamper-proof logging system. The most effective evidence for post-event analysis comes from logging and auditing capabilities, which are essential for mitigating technical issues and providing forensic evidence to detect data tampering or unauthorized breaches. These systems must guarantee the integrity, confidentiality, and availability of data while enabling incident detection, compliance, and forensic investigation \cite{slais2023}.

However, covering tracks in log systems has become common practice for attackers, erasing any traces that could lead to their identification or assisting in the analysis of their actions. Therefore, it is necessary to strengthen the system properties to prevent unauthorized modifications \cite{Ahmad2019}. This includes safeguarding the integrity of log data and implementing measures to detect and counteract tampering attempts. Effective accountability solutions must ensure the detection and isolation of faulty behaviors and their origin. Moreover, they should guarantee search accessibility, enabling access to specific records whose integrity can be verified \cite{chui2021}.

Integrating accountability solutions into robotic systems often demands a trade-off with system performance. Specifically, the computational power and networking bandwidth may be adversely affected \cite{fernandez2021}. The inclusion of anti-tampering techniques, such as full disk and communication encryption, can be restrictive due to resource limitations \cite{Fernandez2024}. Additionally, high-cost tamper-proof storage devices may prove economically unviable, particularly when dealing with continuously generated log data.

Therefore, it is necessary to establish an efficient and verifiable proof of integrity to guarantee the reproducibility and identification of the events that could arise during an autonomous agent action without compromising the robot's performance. For this evidence to be significant, it should belong to a persistent chain of custody, fully reliable and verifiable, key requirements for an auditable and transparent logging system \cite{Putz2019}.

To this end, blockchain solutions have emerged as a mechanism to ensure data integrity, protection against unauthorized changes, and traceability. Consensus protocols established by all blockchain network members enable the detection and rejection of any destruction or manipulation of data through illicit channels. The nature of this technology mitigates tampering risk and safeguards data \cite{iqbal2019comparison}.

Nevertheless, while blockchain secures log data, the evidence obtained poses significant challenges for analysis, due to its diverse and varied nature, terminology, and the presence of substantial volumes of similar messages unrelated to anomalous events. These factors may result in limited usefulness and misalignment with the expectations of non-expert users \cite{Shehu2022}. The raw log messages produced by autonomous agents are typically noisy and semi-structured. In addition, it is necessary to deal with specific features such as large quantities of domain-specific language messages \cite{Ott2021}. Addressing this issue requires methods to organize and filter log data, making it accessible and meaningful.

In this context, explaining the reasons behind a robot's specific and unexpected actions is essential for diagnosing failures and comprehending deviations from the intended goals. Equally important is ensuring that these explanations are understandable to users. Explainable Artificial Intelligence (XAI) is a research area dedicated to enhancing the transparency and interpretability of Artificial Intelligence (AI) systems to ensure their responsible and ethical deployment. As autonomous robots grow in complexity, understanding and predicting their behavior has become increasingly challenging for non-expert users. This requirement has led to the development of eXplainable Autonomous Robots (XAR), which aims to provide mechanisms for presenting clear and understandable explanations for the actions and decisions of autonomous agents \cite{PapagniKoeszegi2021}. XAR enhances the transparency and interpretability of these agents' decision-making processes, often incorporating techniques such as natural language explanations to improve clarity and user comprehension \cite{Anjomshoae2019,Langley}.

To further improve the effectiveness of XAR systems, Large Language Models (LLMs) are used in diverse natural language processing tasks, including command translation and task planning \cite{WANG202552}. Our approach integrates LLMs with blockchain-secured log data. Using retrieval-based techniques, this solution generates accurate and context-aware natural language explanations based on verifiable information \cite{ren2023investigating}. This integration ensures that explanations are clear, coherent, and grounded in a traceable and reliable context, effectively addressing user comprehension and data integrity concerns.

Although recent studies have advanced accountability and explainability in autonomous robotic systems, many existing approaches still face important limitations. Several rely on fixed templates or static rule-based methods that lack flexibility to adapt to dynamic and unpredictable environments. Consequently, the explanations produced may be generic, rigid, or poorly aligned with the actual context of the robot's behavior. Moreover, explainability techniques are often developed separately from accountability mechanisms, leading to explanations not grounded in verifiable data. This disconnect limits their reliability and effectiveness, particularly in safety-critical settings where trust and traceability are essential.

In addition, current evaluation methods for these explanations face significant challenges, including the absence of standardized and domain-specific metrics, biases in the generated text, and the high cost of human evaluations \cite{chang2023surveyEv}. Even though human evaluations remain the gold standard for capturing subjective insights \cite{MILLER20191}, they are resource-intensive and lack scalability \cite{2024Miller}. Alternatively, LLM-based evaluation methods offer a scalable alternative, demonstrating over 80\% alignment with human judgments while offering interpretable and explainable assessments \cite{zheng2023judging,nguyen2024interpretable}.

\subsection{Contributions}

This work presents an accountability and explainability architecture designed for Robot Operating System (ROS)-based mobile robots. This solution integrates two main components. The first is a black box-like module that provides accountability through anti-tampering features by distributing integrity proofs from the events recorded by the black box. The second component generates natural language explanations based on black box data, including interactive natural language conversations as a means of providing comprehensive explainability for the recorded events. Both components are connected through an additional module responsible for processing the raw data collected by the first module, eliminating non-essential information, and building connections and relations between data. These functionalities release software developers from the need to make calls to the logging Application Programming Interface (API) to build explanations to document the code and behaviors of the autonomous agent.

To validate our approach, we conducted a comprehensive evaluation to examine both the impact of the architecture on system performance and the quality of the generated explanations. In order to capture detailed insights while maintaining efficiency and reproducibility, we developed a hybrid evaluation methodology that combines human judgments with LLM-based assessments. The evaluation focused on key criteria such as correctness, accuracy, helpfulness, coherence, and user satisfaction. This integrated framework provides a robust, scalable, and reliable means of assessing LLM-generated explanations in robotic systems. Our findings demonstrate that the proposed architecture successfully balances accountability and explainability without compromising robotic performance, thereby offering a practical and scalable solution for deployment in real-world applications.

Therefore, the primary contributions of this work are summarized as follows:
\begin{enumerate}
    \item A unified architecture that combines blockchain-based tamper-proof logging with LLMs to generate grounded natural language explanations of robot behavior, enabling accountability and transparency.
    \item Real-time integrity and verifiability of robotic event data through a blockchain-based black box component, with minimal impact on robot performance, even at high data rates, addressing a critical requirement for practical deployment.
    \item A log curation and processing module that filters noisy, semi-structured robotic logs and extracts meaningful context, significantly improving the relevance, structure, and coherence of the generated explanations.
    \item An empirical analysis of the trade-offs between system performance and logging granularity, offering practical design insights for developing robust, interpretable, and efficient autonomous agents
    \item A hybrid and comprehensive evaluation methodology combining LLM-as-judge and human-based assessments, designed to leverage the strengths of both approaches in validating the quality, accuracy, and user satisfaction of the generated explanations for complex robotic behaviors such as goal cancellations, re-planning, and obstacle avoidance.
\end{enumerate}


\subsection{Structure of the Paper}

The remainder of this paper is organized as follows: Section \ref{sec:related_works} provides an overview of previous studies and findings related to the study, focusing on the enhancement of immutable data recorders and explainability solutions for robots. Section \ref{sec:system_architecture} delves into the system architecture, offering a detailed description of the designed approaches. Section \ref{sec:implementation} describes the technical elements of the system designed for the proposed architecture. Section \ref{sec:evaluation} presents the evaluation process, including the scenarios and assessment variables. Section \ref{sec:experimental_results} outlines the results obtained, particularly addressing accountability and explainability metrics. Finally, Section \ref{sec:conclusions} summarizes the obtained conclusions.

\section{Related Works}\label{sec:related_works}
In this section, we explore key works related to various domains relevant to our approach. This includes advances in how blockchain ensures the immutability of logs, research on applying Event Data Recorders (EDRs) in robotics, significant contributions to integrating XAI with blockchain, and XAR. Finally, we examine the evolution of LLMs to generate natural language explanations and their advancements in robotics.

\subsection{Blockchain for Immutable Logs}
Blockchain is a decentralized solution that eliminates the necessity for trust in a central authority through its immutable and distributed ledger, composed of timestamped transaction blocks. These blocks are linked through the hashing and storage of the previous block content, ensuring the integrity of transactions. Any attempt to alter a transaction in one block requires modification of all subsequent blocks, resulting in a computationally expensive process. As a result, the data stored in the blockchain is immutable \cite{Rajasekaran2022,Guo2022}.

There are two primary types of blockchains based on access controls: permissionless blockchains allow open access and public transaction visibility, whereas permissioned blockchains limit participation and transaction visibility. Nevertheless, closed networks may pose security risks, as authorized users could recalculate subsequent blocks in the event of block modification \cite{Solat2021}. While permissionless blockchains have traditionally been related to scalability and throughput challenges, approaches such as Ethereum have transitioned to Proof of Stake (PoS), a more energy-efficient, secure, accessible, and scalable consensus protocol \cite{Nguyen2019,SANKA2021103232}.

Blockchain technology ensures log integrity and auditability, offering unique advantages over alternatives such as distributed storage systems or InterPlanetary File Systems (IPFSs), which may require central authorities for storage and verification \cite{Malhotra2021}. Blockchain offers consensus and immutability, coupled with the capability of integrating logic mechanisms such as Smart Contracts (SCs). SCs contain functions that either write values onto the blockchain, thus altering its state, or reading data from it \cite{8494045}. These features make blockchain the preferred choice for applications in which data integrity and trust are required \cite{Nyaletey2019}.

Different secure logging systems rely on permissioned blockchain frameworks for log immutability \cite{shekhtman2019}, introducing accountability even in embedded systems \cite{chui2021}. To achieve a balance between throughput and cost-effectiveness, some works combine both categories \cite{Zhao2023}, developing approaches that include a permissioned blockchain to ensure per-entry immutability and a public non-repudiation solution \cite{Pourmajidi2023}. Nevertheless, inherent features of permissionless blockchains such as transparency and trust, have boosted the development of secure logging systems \cite{Ge2019}.

While previous work often records cryptographic hashes on the blockchain to ensure log integrity and employs SCs for data management, our approach goes further by using these verified records as contextual input for generating natural language explanations of autonomous robot behavior. Additionally, our system includes features that enable public accountability and prevent denial of recorded actions. This design provides strong guarantees against data manipulation and ensures that the explanations produced by our system are always based on trustworthy, tamper-proof information. By bridging immutable records with interpretable explanations, our method enhances transparency and clarity in communicating robot actions.

\subsection{Blockchain for Event Data Recorders in Robotics}
EDRs, or black boxes, are essential for capturing and analyzing events in robotic systems. These recorders help comprehend system behavior, detect anomalies, and ensure safety \cite{Winfield2021,winfield2022ethical}. Therefore, these solutions are required for forensic investigations and ethical assessments of autonomous agents \cite{taurer2018}. The integration of black box modules to enhance robots' introspection capabilities contributes to improving system transparency by attributing responsibilities \cite{fernandez2022,Fernandez2024}. These solutions typically include hashing, signing, and encryption \cite{Schmidt2022}, while ensuring event integrity through anti-tamper mechanisms \cite{SRINIVASADITYA2021103245}.

The blockchain technology in EDRs ensures the integrity of a transparent and auditable record of events. These features enhance accountability and trust in the recorded data by preventing tampering \cite{Salah2019}. Decentralized and distributed storage further eliminates reliance on central authorities and mitigates data loss risks.

Further research explores the application of blockchain technology to improve robotic capabilities, particularly event data recording and management. The Black Block Recorder (BBR) system uses permissioned blockchain for immutable logging, combining Rosbag2 and Distributed Ledger Technologies (DLTs) to ensure data integrity \cite{White2019}. Blockchain applications in robot swarms, including secure communication, data logging, and consensus agreements, are discussed in \cite{strobel2020blockchain}. In \cite{Lopes2021},  a time-segmented consortium blockchain for robotic event registration is proposed, enabling efficient event organization and retrieval within specific time frames. In \cite{Zhang2022}, the authors present a tool to bridge ROS and Ethereum blockchain, evaluating encryption efficiency and stability, transaction response time, and packet loss rate. 


Existing blockchain-based logging solutions in robotics usually focus on low-level data storage and network performance metrics, and often lack mechanisms to record integrity proofs of structured and filtered messages at configurable intervals. Furthermore, these solutions generally miss granular recording policies and the use of SCs to improve data verifiability. In contrast, our approach guarantees tamper-proof evidence by periodically anchoring integrity proofs of curated log data on the blockchain, while also generating natural language explanations grounded in verifiable and filtered information. This integration significantly improves both the accountability of robotic systems and the comprehensibility of the explanations provided to users in dynamic contexts.

\subsection{Explainable Artificial Intelligence and Blockchain}

XAI aims to align AI decisions with human cognitive expectations by employing techniques such as contextual relationships, contrastive explanations, counterfactual reasoning, and interpretable models that enhance interpretability and transparency \cite{Madumal2020, Gunning2019}. Insights from social sciences show that people tend to interpret explanations more positively when they reflect human-like reasoning, emphasizing intentions, goals, and contextual factors rather than purely technical or statistical descriptions \cite{MILLER20191}. By combining XAI with transparent blockchain record-keeping, explanations can be retraced and validated, ensuring both interpretability and accountability in AI-driven systems \cite{nassar2020blockchain, HAN2023100598}.

The benefits of this integration are remarkable in several domains. In healthcare, XAI and blockchain have been combined for secure, transparent patient data management, logical disease prediction, and immutable medical diagnostic records  \cite{JOVANOVIC2024110303}. Similarly, in legal domain applications, blockchain and XAI support tamper-proof decision auditing, using IPFS and Ethereum for cryptographic proofs that enable data verification and accountability \cite{SACHAN2024107666, Malhotra2021}. Privacy-preserving systems with homomorphic encryption and semantic ontologies further ensure data protection, while maintaining explainability \cite{Demertzis2023}.

Regarding autonomous agents, such as unmanned aerial vehicles or autonomous vehicles, blockchain and XAI ensure secure and transparent decision-making \cite{Bendiab2023}. This integration enhances collaboration, even in complex and time-critical scenarios \cite{Calvaresi2019}. Decentralized edge models for collaborative AI leverage blockchain for secure communication and real-time data sharing, further optimizing resources, and enhancing reliability in these environments \cite{Shehata2021}.

Despite previous advances, providing explanations grounded in robot event data remains challenging due to the complexity, volume, and heterogeneity of the generated information. To address this, our proposal introduces Rosbag2 files as an alternative to IPFS or distributed file systems for storing contextual data in autonomous robotic systems. This approach supports reproducibility and enables a more precise analysis of robotic behaviors \cite{White2019}. Rosbag files provide a standardized format for recording large volumes of robotic data, facilitating comprehensive interaction capture for improved analysis. To mitigate the inherent indeterminism of LLM algorithms, we propose using a permissionless blockchain to store integrity proofs of Rosbag messages, ensuring data immutability, simplifying verification, and increasing reliability in explanation generation. 

\subsection{Explainable Autonomous Robots}
XARs are designed to provide explanations of their actions and decisions to non-expert users. Such explanations enhance security and trust, helping to prevent failures, errors, conflicts, and undesired interactions while achieving objectives \cite{Han2019,Sakai2023}. However, despite the growing demand for explainability in autonomous agents, their use lacks practical usability and interpretability in real environments \cite{Abdul2018}. 

Effective explanations in XARs should detail the alternatives, choices made, expectations for each option, decision criteria, and any unexpected events that may have led to changes in the planned behavior \cite{Langley, Sadeghi2021}. These explanations often rely on abductive reasoning, which bridges the gap between the robot’s internal decision-making process and the user's understanding \cite{Sakai2022}. To support this, frameworks have been developed that enable robots to articulate their internal states and decisions, enhance collaboration and improve communication with humans \cite{sheh2017did}.

Adaptive algorithms and needs-based motivational systems allow robots to provide real-time, context-aware explanations tailored to diverse audiences \cite{stange2022self, SETCHI20203057}. Behavior Trees (BTs) hierarchically decompose tasks into goals and subgoals, making complex behaviors easier to understand and adaptable to real-time changes \cite{Han2021}. Nevertheless, achieving a balance between simplicity and adequacy remains a key challenge when conveying complex or technical information in an understandable way. Explanations that are overly detailed can overwhelm users, while those that are too simplified may omit critical context necessary for constructing accurate and meaningful interpretations \cite{MUALLA2022103573}.

Our architecture addresses these limitations by incorporating a log curation module that automatically processes and filters raw robotic data, retaining only the most relevant information for explanation generation. By emphasizing essential details and excluding non-essential information, our system produces explanations that align with human cognitive capacities, thereby fostering trust and facilitating effective collaboration.

In addition to providing context-aware explanations, effective explainability in XAR fundamentally relies on clear communication \cite{Sakai2022}. While prior research has acknowledged the potential of Natural Language Processing (NLP) and LLMs to enhance human-robot communication \cite{Sakai2022, Sado2023}, their integration with robot-specific data streams remains limited. This limitation arises because LLMs primarily depend on probabilistic patterns rather than direct access to data from the robot, which can lead to responses that are linguistically plausible yet factually inaccurate. Our approach addresses this challenge by grounding explanations in curated robotic data and leveraging the natural language capabilities of LLMs, thereby producing clear and accessible explanations that enhance transparency and foster trust between users and autonomous systems, including non-expert audiences.

\subsection{Large Language Models in Robotics}
LLMs, enabled by self-attention mechanisms, have advanced NLP, improving task-solving capabilities and enabling more effective human-robot interaction \cite{chang2023survey, zeng2023largelanguagemodelsrobotics}. However, LLMs face challenges such as hallucinations, which arise due to reliance on internal knowledge, and the high update costs associated with large parameter sizes, making traditional pre-training and fine-tuning methods impractical. Retrieval-Augmented Generation (RAG) methods address these issues by enabling LLMs to interact effectively with external sources \cite{andriopoulos2023augmenting, zhao2023retrieving}. This approach is particularly advantageous in robotics, where real-time and accurate contextual understanding is essential for complex decision-making and task execution.

Explainability and transparency have become critical for deploying LLMs in autonomous agents. These models enable robots to translate high-level commands into actionable plans, improving their performance in complex environments \cite{Wilcock2023, wang2024largelanguagemodelsrobotics, meng2024llmalargelanguagemodel}. When integrated with reward decomposition frameworks, LLMs can generate intuitive explanations, linking actions to object-specific properties, thereby reducing ambiguity \cite{Lu2023}. LLMs can also generate natural language descriptions of robot behaviors and logs, highlighting opportunities for improvement in accuracy and completeness \cite{setianto2022, gonzalez2023using}.

However, existing solutions often rely on static models or predefined data, which limit their ability to handle dynamic environments. Furthermore, these solutions rarely provide mechanisms for ground explanations in verifiable, tamper-proof evidence of robot behaviors. Our approach addresses these limitations by integrating blockchain-supported accountability with the curated logging of relevant robotic data. This framework provides a robust context for LLM-generated explanations, minimizes hallucinations, and ensures alignment with the robot's actions and interactions within its environment.


\section{System Architecture}
\label{sec:system_architecture}
Our approach comprises two main components: a tamper-proof black box-like device, and a module responsible for generating natural language explanations for the data stored by the former. The first component offers services for building, storing, and verifying integrity proofs through the SC specifications. The explainability component provides natural language explanations by implementing RAG, thereby enhancing the results obtained using the data recorded by the first component as an external source to enrich the context in the answer formulation process.
Both components are integrated through the existence of two main asynchronous tasks focused on cleaning and processing the raw data generated by the black box-like device to increase the effectiveness of the explainability engine.

\begin{figure*}
	\centering
	\includegraphics[height=9.0cm]{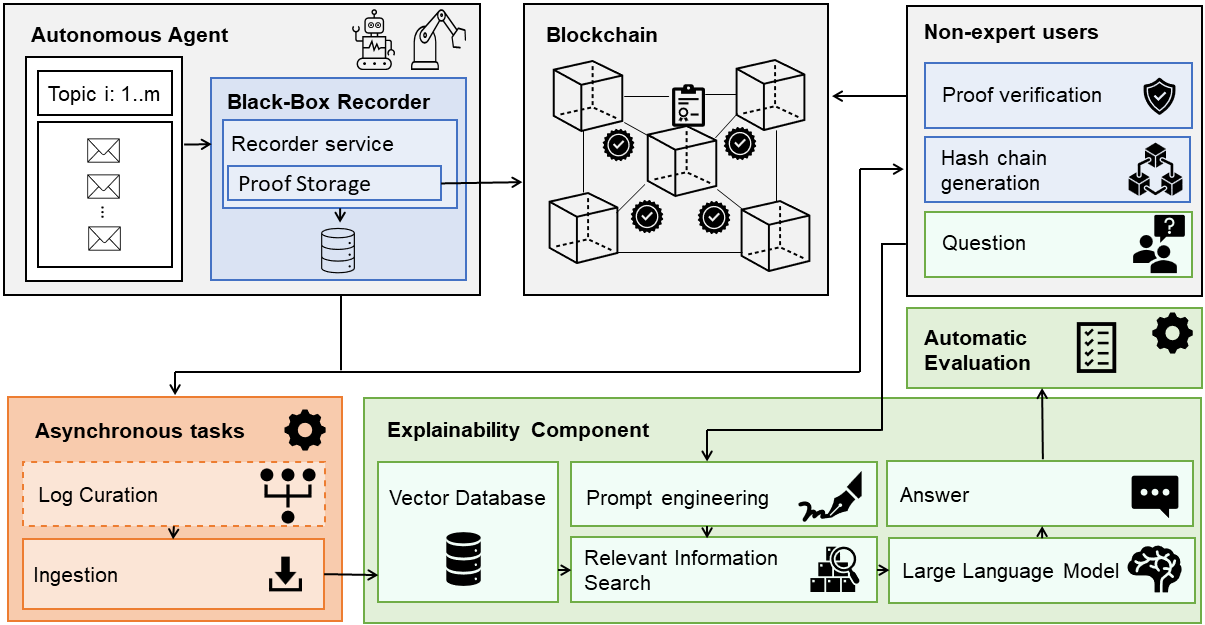}
	\caption{System architecture.}
	\label{fig:sytem_architecture}
\end{figure*}

Figure \ref{fig:sytem_architecture} depicts the generic architecture of our proposal, highlighting the functionalities related to accountability in blue, and those related to explainability in green. Below, we provide a deeper description of the approaches designed and followed to develop and integrate the above-mentioned components.

\subsection{Black Box Recorder}\label{sec:BBR}
The first component focuses on recording accountable information that captures the essential messages generated during the navigation task of an autonomous agent. The selected data will guarantee the reproducibility of the mentioned task, including details such as the robot's poses, employed map, odometry data, linear and angular velocity, and laser scan readings.

The immutability of the previously described information is achieved using blockchain. Blockchain technology usually faces limitations in terms of the size of the data that can be written into a single transaction. Efficient storage of autonomous agent messages is challenging owing to these constraints. Our proposed solution involves selectively storing the hashes of messages at intervals, providing the chance to use different storage frequencies depending on the topic and its relevance. This fact not only enhances the granularity of stored message proofs but also provides a more adaptable solution.

Our anti-tampering approach to this accountability proposal comprises two key components: a chain of integrity proofs coupled with an SC specification. This section delves into their design and explains how both align with the constraints of the autonomous agents.

Key features such as consistency are guaranteed by blockchain's consensus protocol, in which the content of each block is hashed and stored in the next one. In addition, availability is assured by the distributed and decentralized nature of the blockchain, ensuring accessibility across the entire network.

\subsubsection{Chain of Integrity Proofs}
To preserve the integrity of the messages without compromising system performance, our solution entails building a chain where each message hash is linked with the previous hash value. This approach ensures that any alteration of the selected messages results in a different final hash, easing tampering detection. The verification process is also simplified because only the last hash in the chain must be recomputed and compared with the last stored value. A single hash modification disrupts the entire chain from that point onward, simplifying the identification of tampering.

We define a chain entry ($CE$) with a counter $i \in 1..m$  to be linked with the previous one by concatenating the prior digest $h_{i-1}$ with the current log message $Msg_i$. \textit{hash()} denotes the cryptographic hash function, with SHA-256 being the chosen one, which is commonly used for data integrity verification and digital signatures. 


The initial hash value $h_0$ of the chain is calculated by applying the selected hash function to a set of $l=32$ random bytes ($RB$), to generate a 256-bit value. Introducing initial entropy into the chain of integrity proofs enhances privacy,  ensuring that the integrity proofs for similar initial log messages from different records are not identical. Data privacy is further prioritized by storing only the hashes in the blockchain, rather than the content of the messages.

\begin{equation}
\begin{aligned}
  CE_i &= h_i \\
  \text{where}~ RB &\leftarrow \{0..F\}^l,  \\
  h_0 &\leftarrow \text{hash}(RB), \\
  h_i &= \text{hash} (h_{i-1} || Msg_i)
\end{aligned}
\label{eq:chain_entry}
\end{equation}

By chaining every hash with the previous one, we optimize the validation of the stored messages because the last linked digest should correspond to the final proof saved in the blockchain. Additionally, a service has been developed to generate the chain of hashes using the recorded data, ensuring partial validation of the integrity proofs without modifying the stored messages or adding non-essential information to the transactions saved in the blockchain.

The number of proofs included in a blockchain transaction depends on the transaction cost limit set by the blockchain framework. Therefore, being $j \in 1..m$ a transaction counter, each newly arriving chain entry is added to a transaction payload $P_j$ until the limit is reached. Being \textit{sign()} a shorthand for a signature function and $PrivKey$ the private key of the account, transaction $T_j$ is transmitted from the service in charge of recording accountable data to the blockchain.

\begin{equation}
\begin{aligned}
  T_j &= (P_j, S_j) \\
  \text{where}~ P_j &\leftarrow \{CE_i, CE_{i+1}, CE_{i+2}, \dots\}, \\
  S_j &= \text{sign}(P_j, PrivKey)
\end{aligned}
\label{eq:bc_transaction}
\end{equation}

A service interruption or a disruption in the behavior of the autonomous agent will lead to the aggregation of the remaining hash values into a single transaction, even if there is capacity for additional values. This practice guarantees the immutability of the last messages sent by the robot upon completing a task or achieving a goal.

\subsubsection{Smart Contract}
Calling an SC function intended for blockchain writing involves the execution of a transaction that changes the state of storage data on the blockchain. In our approach, this transaction is signed with the account's private key via the Elliptic Curve Digital Signature Algorithm, effectively identifying the signer.

The defined SC also determines its ownership, implementing access control to ensure that only the contract owner can add or update message proofs. These features guarantee authentication and non-repudiation.

Moreover, our SC includes a function to verify the existence of a message proof on the blockchain. This allows external parties to independently confirm whether a specific message hash has been stored in the contract and when it is stored. This functionality contributes to the transparency of the system, enabling users or auditors to check the validity of message proofs without relying solely on the contract owner, thereby promoting public accountability.

\subsection{Explainability Component}\label{sec:explainability_component}

Effective communication is essential for achieving explainability in autonomous robots. Our approach focuses on improving this feature through the use of NLP technology, enabling non-expert users to interact with agents in a meaningful way.

Our solution involves the use of RAG to retrieve data from outside the language model (non-parametric) and augment the prompts by adding the information recorded from our accountability component in context. Before this integration, the information is processed by a dedicated module responsible for acquiring and processing the raw data generated by the autonomous agent, which serves as contextual input in the question-answering system. 

Given the volume of messages, their varied nature, their length, the presence of non-meaningful records generated by the autonomous agent, and the need to establish specific relationships between messages to enhance the context of the engine, the log curation component extracts the relevant information. This functionality enhances the effectiveness of the explanations provided to non-expert users. A more comprehensive description of this component is provided in Section \ref{sec:implementation}.

After processing the raw data stored by our accountability component, the ingestion process performs a chunking task to break down extensive text segments into more manageable units. To ensure comparability during relevancy searches at inference time, both the data and user query are converted into numerical representations by embedding language models. Next, based on the embedding of the user query, the relevant text is identified in the document collection using a similarity search in the embedding space. The user-provided prompt is then extended with the relevant information obtained in the search and added to the context. This enhanced prompt is sent to the LLM. Because the context contains pertinent external data along with the original prompt, the model output is relevant and accurate.

The addition of new accountable data to our explainability component requires asynchronous updates of accountable data and its embedding representation. These updates ensure that the model is sensitive to changes, thereby addressing the limitations introduced by offline training.


\section{Implementation} \label{sec:implementation}
The solution presented in Section \ref{sec:system_architecture} can be implemented using various technologies. This section details the specific solutions used to develop our proposal.

\subsection{Black Box Recorder}

ROS is the most widely used framework in robotics, offering a collection of software libraries and tools for developing robot applications. Nevertheless, limitations in fulfilling the requirements of real-time embedded systems and the necessity to enhance multi-platform support led to a major upgrade to ROS 2, whose main change lies in the adoption of the Data Distribution Service (DDS)~\cite{Maruyama2016}.

Nodes constitute fundamental components of both ROS and ROS 2 applications. These nodes are individual processes that handle specific tasks and communicate with one another through messages. Data transfer between nodes can be achieved using various methods, such as topics, services, and actions. Topics facilitate message passing, services enable synchronous request-reply interactions, and actions are utilized for long-running tasks that require feedback.

This study focuses on navigation capabilities. For this purpose, we utilized Nav2, a suite of tools for ROS 2 that is regarded as the successor of the ROS Navigation Stack. Nav2 equips the robot with the capacity to execute complex navigation tasks.

Nav2's core functionalities include planning, control, localization, visualization, and additional features. Furthermore, it employs BTs to create customized navigation behavior by orchestrating multiple independent modular servers, offering high flexibility in navigation tasks and specifications of complex robot behaviors \cite{macenski2020marathon2}. Different independent task servers, intended to compute a path, control effort, or recovery, communicate with the BT over a ROS 2 interface, such as an action server or service. The use of BTs to generate explanations for robot behavior has previously been successfully evaluated, allowing causal information to answer questions \cite{zhao2021explainabilityBTs}.

In ROS 2, nodes communicate by publishing data on topics that can be consumed by other nodes. The data can range from sensor readings that provide the robot's perception of its environment to control commands sent to actuators from decision-making nodes. The navigation reproducibility of autonomous robots in ROS 2 involves storing essential data such as their poses, maps, odometry, velocity, and laser scan readings. To this end, we used Rosbag files, designed to capture and store data generated during the execution of a ROS system. These files can reproduce the exact conditions and inputs from a previous run, making them particularly valuable for understanding the system behavior. The set of topics included in the generated Rosbag files generated are detailed in Table \ref{tbl1}. The selected topics capture the core aspects of the robot’s operation, including localization, environment perception, navigation status, decision-making logic, and motion control. This selection ensures that the recorded data provides a coherent and sufficient basis for interpreting and reproducing the robot’s navigation behavior.

\begin{table*}[h] 
    \caption{Topics name included in the Rosbag file their description.}\label{tbl1}
    
    \begin{tabular*}{\linewidth}{ @{\extracolsep{\fill}}l l}
        \toprule
        Topic name 1 & Information provided \\
        \midrule
map  &   2-D grid map, in which each cell represents the probability of occupancy. \\
tf\_static  &   Static transform information; fixed relationships between coordinate frames over time.\\
robot\_description  & URDF (Unified Robot Description Format) or robot model description. 
\\
navigate\_to\_pose/\_action/status  &  Information about the current state or progress of the navigation action. \\
global\_costmap/costmap & Representation of the environment in terms of costs associated with different areas.\\
plan  & Representation of the planned path for the robot to follow  as a sequence of waypoints.\\
rosout & Logs and messages from ROS nodes. \\
local\_costmap/costmap  & Representation of the immediate surroundings of the robot and their cost.\\
amcl\_pose  & Position and orientation of the robot. \\
behavior\_tree\_log & Nav2 Behavior Tree nodes status.\\
cmd\_vel  & Linear and angular velocity commands to the robot's motion control system. \\
camera/image\_raw  & Raw image data captured by a camera sensor on the robot. \\
scan  & Distances from the sensor to obstacles in the robot's surroundings. \\
odom  & Estimated motion and position of a robot based on sensor data.\\
tf  & Coordinate frame transformations in the robotic system. \\
        \bottomrule
    \end{tabular*}
\end{table*}

Ethereum has been selected as the blockchain framework to ensure the immutability of generated Rosbag files. Ethereum, an open-source public blockchain, introduces the cryptocurrency Ether and self-executing programs (SCs) for code execution on the blockchain. Ethereum's open-source nature ensures decentralization and accessibility, allowing participation without requiring approval, which is crucial for maintaining the integrity of the recorded data. Additionally, Ethereum's use of PoS as a consensus protocol enhances energy efficiency and throughput compared to traditional protocols such as the Proof of Work. Furthermore, Ethereum's scalability, security and economic efficiency make it a secure option for ensuring data immutability. High-level Algorithms \ref{alg:accountable_bc_recording} and \ref{alg:proof_verification} depict the main steps in the recording and proof verification services.

 \begin{algorithm}
\caption{Accountable Information Recording}
\label{alg:accountable_bc_recording}
\begin{algorithmic}[1]
\REQUIRE $topicsSet$ = [$selectedTopics$], $serviceCallValue$, $contractAddress$, $contractAbi$, BLOCKS\_LIMIT, $privateKey$
\ENSURE $rosbag$

\STATE $recording \leftarrow$ false
\STATE $initialNonce \leftarrow$ sha256($randomBytes$)
\STATE $recording \leftarrow serviceCallValue$
\STATE $proof \leftarrow \{\}$
\IF{$recording = true$}
  \STATE $previousHash \leftarrow initialNonce$
  \FOR{$message$ \textbf{in} $topicsSet$}
    \IF{$messageNumber$ \textbf{mod} $topicRate$ = 0}
      \STATE $chainedHash \leftarrow$ sha256($message$, $previousHash$)
      \STATE $previousHash \leftarrow chainedHash$
    \ENDIF
    \IF{$payloadSize$ < BLOCKS\_LIMIT}
      \STATE $payload \leftarrow payload \cup chainedHash$
    \ELSE
      \STATE loadBcContract($contractAddress$, $contractAbi$)
      \STATE $transaction \leftarrow$ buildTransaction($payload$)
      \STATE $signedTx \leftarrow$ signTransaction($transaction$, $privateKey$)
      \STATE sendTransaction($signedTx$)
      \FOR{$hash$ \textbf{in} $payload$}
        \STATE saveBC($proof[hash]$ = BLOCK\_NUMBER)
      \ENDFOR
    \ENDIF
  \ENDFOR
\ENDIF

\RETURN rosbag
\end{algorithmic}
\end{algorithm}

\begin{algorithm}
\caption{Proof Verification}
\label{alg:proof_verification}
\begin{algorithmic}[1]
\REQUIRE $hashValue$, $contractAddress$, $contractAbi$, $privateKey$
\ENSURE $msg$

\STATE loadBcContract($contractAddress$, $contractAbi$)
\STATE $transaction \leftarrow$ buildTransaction($hashValue$)
\STATE $signedTx \leftarrow$ signTransaction($transaction$, $privateKey$)
\STATE sendTransaction($signedTx$)

\STATE $blockNumber \leftarrow$ readBC($hashValue$)
\IF{$blockNumber \neq 0$}
  \STATE $msg \leftarrow$ ``The hash value is stored in block '' $+ blockNumber$
\ELSE
  \STATE $msg \leftarrow$ ``The hash value is not stored.''
\ENDIF

\RETURN $msg$
\end{algorithmic}
\end{algorithm}

Our approach addresses challenges such as the drop in robot performance during the Rosbag recording process and the costs associated with storing blockchain data. We accomplish this by saving chained hashes from messages at specific intervals, ensuring the integrity of the data while preserving system performance. These intervals define how often an integrity proof is stored and can be established based on the relevance of messages from a topic, regarding their required immutability features. Furthermore, message proofs are stored in real-time in the Ethereum blockchain, eliminating the need to modify the content of the Rosbag file. The detection of any alteration to the messages within this file is facilitated by a service that generates a hash chain from the file at intervals set for each topic during the recording process. Blockchain logic has been extended by developing a SC using Solidity language to verify the existence of any of these hashes in the blockchain. Solidity is an open-source programming language that facilitates the creation of decentralized applications, contracts, protocols, and agreements. This contract provides functionalities for storing and verifying proof information for hashes stored on the Ethereum blockchain by associating each hash value with the block number in which it is stored. The verification of transaction ownership is also conducted during the storage of these proofs.

Therefore, this approach combines immutability, decentralization, traceability, and cost efficiency to securely and transparently address challenges related to the real-time recording and storage of Rosbag files.

\subsection{Log Curation Component}
To filter the relevant information from the Rosbag file messages, a raw data-processing node has been implemented. Given the size, variety, and intricate nature of the messages contained in the Rosbag file, the inclusion of this raw information might be challenging for non-expert users. To address this, a dedicated component has been developed to refine the data by discarding non-essential content before generating natural language explanations. Furthermore, by analyzing messages from the recorded Rosbag topics, this service can identify key circumstances in the navigation process, such as the appearance of an obstacle or a change in the planned trajectory.

The ROS 2 \textit{NavigateToPose} action server commands the robot to navigate toward a specific goal. Although Nav2 usually prioritizes the shortest path, external factors may require adjustments. The \textit{/plan} topic provides the sequence of poses planned to reach the goal. If unexpected circumstances cause a deviation from this path, the Euclidean distance between consecutive poses in the precomputed plan will increase. This change, together with sensor data from the \textit{/scan} topic, can indicate potential obstacles that cause re-planning. These reasonings have been successfully verified in previous authors' work \cite{fernandez2023} to build explanations for autonomous agents based on the use of ROS 2 topics.

\begin{figure*}
	\centering
	\includegraphics[height=5cm]{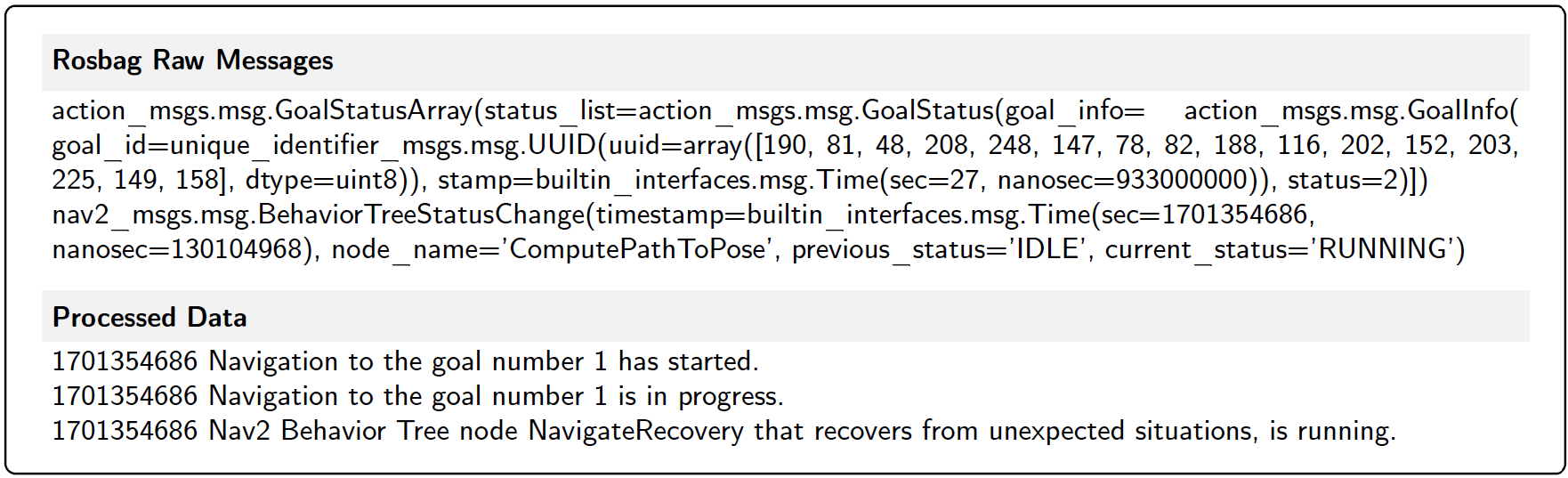}
	\caption{Example of raw Rosbag messages and information obtained from them after the Log Curation process.}
	\label{fig:InterpreterExample}
\end{figure*}

Additionally, the present approach uses the information provided by Nav2 behavior trees because their potential for designing, visualizing, and understanding the behavior of autonomous robots. Other information, such as the present position of the robot and its velocity is also processed through this component. A general overview of its behavior is presented in Algorithm \ref{alg:log_curation}.
As a result, through this component, ROS 2 developers are relieved from calling the logging API to generate explanations. This functionality streamlines the development process, allowing developers to focus on core functionalities without the need for managing log calls. This approach reduces debugging efforts and boosts efficient resource usage, scalability, and adaptability in the development lifecycle, contributing to a more productive natural language explanation generation process. An example of the transformation of ROS messages through this component is shown in Figure \ref{fig:InterpreterExample}.

\begin{algorithm}
\caption{Log Curation Process}
\label{alg:log_curation}
\begin{algorithmic}[1]
\REQUIRE $rosbag$, $btNodeDesc$
\ENSURE $curationLog$

\FOR{$message$ \textbf{in} $rosbag$}
  \STATE $msgTopic \gets \text{getMsgTopic}(message)$
  \STATE $logMsg \gets message.data$
  
  \IF{$msgTopic = \text{``/navigate\_to\_pose/\_action/status''}$}
    \STATE $logMsg \gets$ Track navigation status
  \ELSIF{$msgTopic = \text{``/rosout''}$}
    \STATE $logMsg \gets$ Record log outcomes
  \ELSIF{$msgTopic = \text{``/plan''}$}
    \STATE $logMsg \gets$ Log path changes and obstacles
  \ELSIF{$msgTopic = \text{``/behavior\_tree\_log''}$}
    \STATE $logMsg \gets$ Log behavior tree node status
  \ELSIF{$msgTopic = \text{``/amcl\_pose''}$}
    \STATE $logMsg \gets$ Record robot position and orientation
  \ELSIF{$msgTopic = \text{``/cmd\_vel''}$}
    \STATE $logMsg \gets$ Log linear and angular velocities
  \ENDIF

  \STATE Append time-stamped $logMsg$ to $curationLog$
\ENDFOR

\RETURN $curationLog$
\end{algorithmic}
\end{algorithm}

\subsection{Explainability Component}

\begin{figure*}
	\centering
	\includegraphics[height=8.4cm]{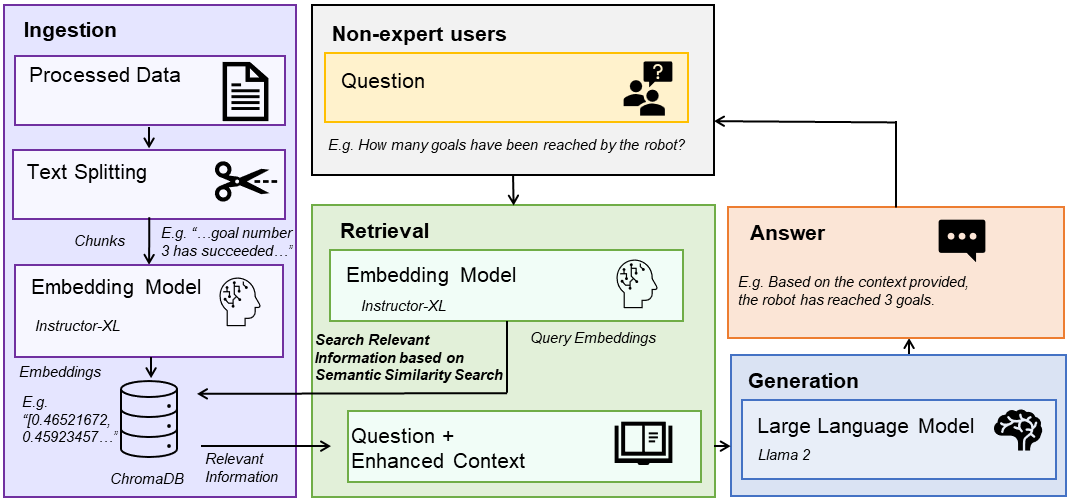}
	\caption{Retrieval Augmented Generation architecture.}
	\label{fig:RAG_architecture}
\end{figure*}

The output of the previously described node serves as an external source in the implementation of the RAG approach. This development is based on the use and extension of localGPT \footnote{https://github.com/PromtEngineer/localGPT}, an open-source question-answering system designed to interact conversationally with documents while safeguarding the privacy of the underlying information. Running locally, it ensures data privacy and secure interactions with the interpreted and recorded data, facilitating the generation of natural language explanations. Its functionality involves loading a set of selected documents, breaking them into smaller chunks, augmenting search capabilities with LangChain tools, and generating local embeddings. The outcomes are stored in a local vector database using Chroma vector storage. Once this information is saved, and given a user input, relevant splits are retrieved from storage via a similarity search, which finds the appropriate piece of information from the documents. The retrieved data is used by the selected LLM when producing an answer. Figure \ref{fig:RAG_architecture} depicts the main components in the RAG architecture used.

Therefore, this implementation combines the strengths of robotics frameworks, blockchain technology, and natural language processing to provide a comprehensive solution for real-time recording, secure storage, and user-friendly interpretation of robotic system data. It addresses challenges related to performance, storage costs, and user comprehension, making it an effective approach for understanding the behavior of autonomous robots.

\subsection{Software and Hardware Artifacts}

Our solution has been deployed on a Linux server running Ubuntu 22.04 as the operating system. The server is equipped with a 6-core 2.6 GHz CPU, 64 GB of memory, and a 4 GB GPU.

The chosen ROS 2 version is Humble. Details regarding the topics incorporated into the Rosbag file, along with their average rates, are provided in Table \ref{table:ROS2Topics}. These topics play a crucial role in ensuring the reproducibility of the navigation process, facilitating data analysis and debugging.

The experimentation was executed entirely in Gazebo, an open-source solution for 3D robotic simulations. To provide a realistic scenario, we used a hospital simulation environment made available by the Amazon Web Services\footnote{https://github.com/aws-robotics/aws-robomaker-hospital-world}. Additionally, we selected RB-1, an autonomous mobile robot based on the ROS 2 platform produced by Robotnik~\cite{Guzman2017}. This combination of tools and platforms enabled us to assess the impact and effectiveness of our approach.

Interaction with Ethereum has been developed through the web3.py library (version 6.6.1). Ganache (version 2.7.1) has been used as the Ethereum network simulator.

RAG solution holds Langchain (version 0.325), and ChromaDB (version 0.4.6). We used the open-source model Llama 2 7B GGUF \cite{touvron2023llama} as the LLM responsible for generating natural language explanations for end users. This LLM integrates Reinforcement Learning from Human Feedback (RLHF) and shows versatility in handling general-purpose tasks, ranging from answering questions, text generation, and language translation to more demanding tasks such as reasoning or coding. Preliminary tests of Llama 2, compared to other LLMs such as Mistral \cite{jiang2023mistral} or Zephyr \cite{tunstall2023zephyr}, led us to choose this model for evaluating our data due to its promising results and suitability for the available hardware during the development phase of this work. As a text embedding model we have chosen Instructor-xl \cite{su2022one}, a high-performance model designed to generate text embeddings tailored to tasks such as classification, clustering, text evaluation or retrieval, which is particularly important in the processes performed for any RAG solution.

\begin{table}[ht!]
\centering
\caption{Rosbag topics rate and average rate per message.}
\begin{tabular}{lcc}
\toprule
 Topic & Avg. Rate (Hz) & Avg. Size Msg. (B)  \\
\midrule
map  &   0 & 1007308.8\\
tf\_static  &  0 & 9011.2\\
robot\_description  & 0 & 25497.6\\
navigate\_to\_pose/\_action/status  & 0.028 & 2389.333\\
global\_costmap\/costmap & 0.500 & 996147.2\\
plan  & 0.884 & 3809.28\\
rosout & 1.014 & 746.057\\
local\_costmap\/costmap  & 1.646 & 3986.99\\
amcl\_pose  & 1.925 & 1054.72\\
behavior\_tree\_log & 2.74 & 313.224\\
cmd\_vel  & 18.151 & 147.110\\
camera/image\_raw  & 25.57 & 1037110.93\\
scan  & 47.173 & 8846.012\\
odom  & 92.946 & 779.618\\
tf  & 387.278 & 195.413\\
\bottomrule
\end{tabular}
\label{table:ROS2Topics}
\end{table}

\section{Evaluation}\label{sec:evaluation}

To evaluate our proposal, we conducted an experimental study to compare the performance of an accountability system built in accordance with Section \ref{sec:implementation}. This study covered several approaches to storing integrity proofs across various scenarios. Additionally, we explored and adapted the capabilities of RAG solutions and LLMs in question-answering systems, using the data gathered by our black box-like component as a source, through a comprehensive set of questions. This study aimed to assess the effectiveness of these technologies in delivering natural language explanations to individuals who are not experts.

The experiments conducted within this proposal focus on navigation tasks included in the Navigation Functionality of the European Robotics League Consumer Service Robots Challenge \cite{ERL}.

This functionality evaluates the capability of a robot to navigate accurately and autonomously within a given scenario, including furniture, objects, and people. Details, such as the initial position and the number of waypoints the robot needs to visit, are provided beforehand. Furthermore, the coordinates of the waypoints are communicated sequentially to the robot during the runtime. Each waypoint is characterized by its $X$, $Y$, and $\Theta$ coordinates. Existing elements in the environment, such as furniture, doors, and walls, must be mapped. Obstacles, with varying shapes and sizes, remain unknown to the teams beforehand and may differ between runs.

Our experiments are structured around three distinct assessment scenarios in line with the navigation functionality previously described. In the initial scenario, the robot must navigate three waypoints, encountering no obstacles along the way. The second scenario presents the robot with the challenge of adjusting its previously planned path to reach the second goal. This adjustment becomes necessary because the appearance of an obstacle obstructing the original planned route, which requires a deviation to achieve the goal. The third scenario introduces a new obstacle that blocks the only door required to reach the first goal. Consequently, this obstacle leads to the cancellation of the first waypoint objective. The second and third goals are achieved with the same restrictions as those detailed in the second scenario.

\begin{figure}
    \centering
    \subfigure{\includegraphics[scale=.30]{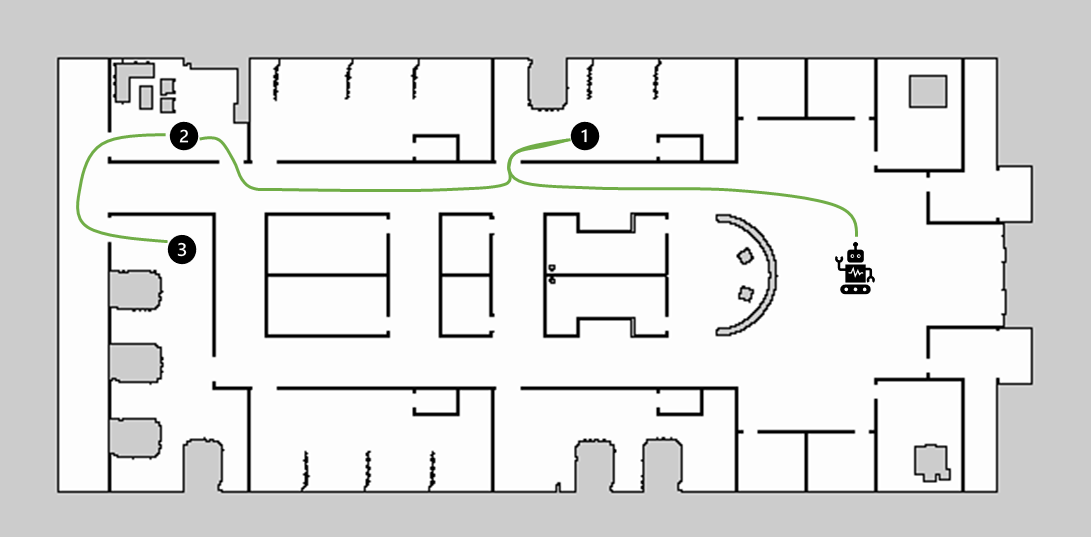}} 
    \subfigure{\includegraphics[scale=.30]{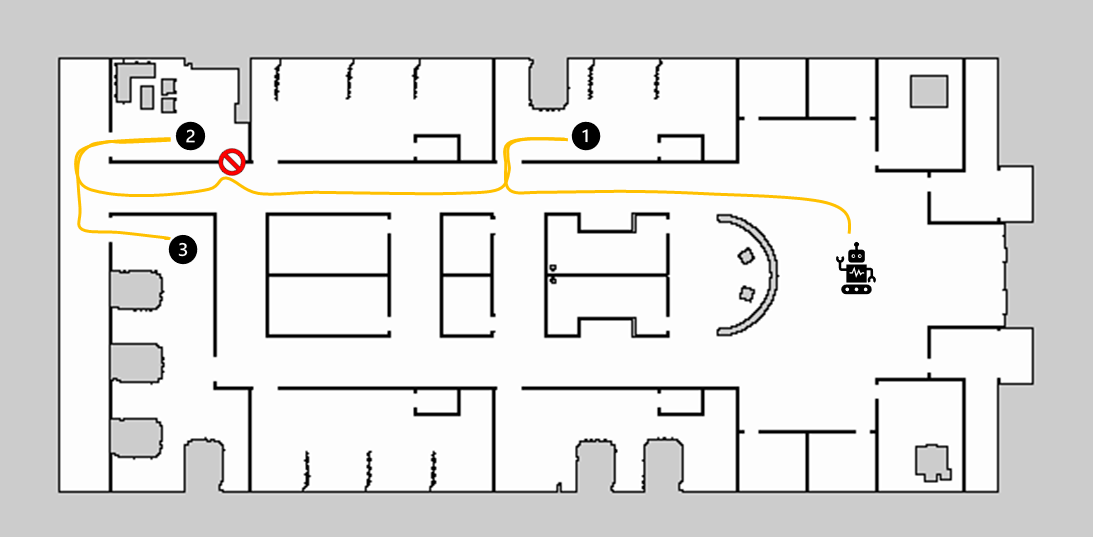}} 
    \subfigure{\includegraphics[scale=.30]{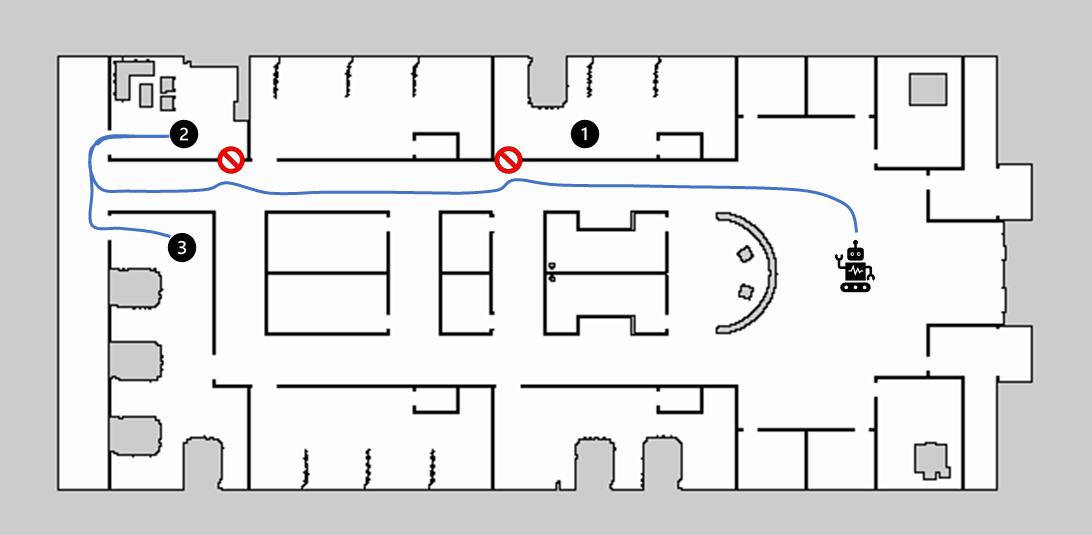}}
    \caption{Scenario maps with trajectories, goals, and obstacles. (a) Scenario 1: No obstacles scenario. (b) Scenario 2: One obstacle scenario. (c) Scenario 3: Two obstacles scenario.}
    \label{fig:scenarios}
\end{figure}

The three scenarios are illustrated in Figure \ref{fig:scenarios}.

The entire source code is accessible online on GitHub\footnote{https://github.com/laurafbec/immutable\_explainable\_BBR.git}, and licensed under the GPLv3.

\subsection{Black Box Recorder Evaluation}

To evaluate the accountability component, we conducted six experiments using the three previously described scenarios. These experiments involved storing integrity proofs on the blockchain, with hash calculations occurring at intervals of one for every 10, 25, 50, or 100 messages. Moreover, our solution introduces the flexibility to save proofs of evidence at different rates based on the topic, ensuring non-tampering for critical topics or those with lower publishing rates where message loss could be crucial.

Therefore, our fifth recording method is designed to store a hash for every message within a topic with a rate below 0.5 Hz, one hash for every 5 messages for topics with rates below 1 Hz, a hash for every 10 messages if the rate is below 3 Hz, one hash for every 15 messages if the topic's rate is below 25 Hz, a hash for every 50 messages for topics with rates below 45 Hz, and one hash for every 100 and 1000 messages when the rate falls below 100 Hz and 400 Hz, respectively. This approach, together with those operating at fixed intervals, has been compared with the generation of a Rosbag file that covers identical topics but lacks anti-tampering properties. Each experiment has been repeated six times, totaling 108 runs, to derive insights into the characteristics of the black box-like device.

\subsubsection{Metrics}

We employed different metrics to evaluate the impact of the accountability system in the chosen scenarios. First, using the topics' publication rate and the number of messages stored per topic in each Rosbag file, we analyzed the average rate of lost messages based on the blockchain storage approach.

Next, regarding system performance, we studied variables such as CPU, memory and disk usage. Specifically, we examined the following variables: system load 1-, 5- and 15-minute average, CPU usage (\%), memory usage (GB), sent network traffic (MB/s) and disk writing rate (MB/s). These variables are highly significant for system-overhead analysis given that elevated load conditions can potentially influence a robot's behavior and event-recording capabilities. In particular, we obtained a full report on system load, CPU, RAM, network traffic, and disk write operations updated every second, facilitating the tracking of overall system activity and an easy comparison with preceding values. To describe the overall distribution and characteristics of these results, we calculated their average as a statistical measure. This information is presented in Section \ref{sec:experimental_results}.

\subsection{Explainability Evaluation}\label{ExplanainabilityEvaluation}
Despite the remarkable performance of LLMs in various language tasks, evaluating these models presents challenges, particularly in ensuring factual consistency. The extensive capabilities of LLMs and the limitations of current benchmarks in capturing human preferences have led to emerging trends in LLM evaluations, such as the use of strong LLMs as judges \cite{zheng2023judging}.

Traditional similarity-based metrics, such as BLEU and ROUGE, have been proven to be misaligned with human evaluations. Although human evaluation remains the primary standard for assessing human preferences, it is exceptionally time-consuming and expensive. Furthermore, this approach significantly affects the scalability of evaluations, making it necessary to use more effective and unbiased methods \cite{2024Miller,hosking2024humanfeedbackgoldstandard}. To automate the evaluation process, several works have adopted the \lq LLM-as-a-judge' approach, which focuses on using strong LLMs, such as GPT-4. These models are often trained with RLHF, resulting in strong human alignment \cite{LIN2024122254}. Previous studies have shown the consistent agreement between these models and human grading \cite{bubeck2023sparks,Leng2023,NEURIPS2023Rafailov}. Moreover, research shows that LLMs can replicate most of conclusions from human evaluations, particularly in tasks involving prediction accuracy and explanation helpfulness, even across varied settings and domains \cite{2024Miller,NEURIPS2023Zhou,NEURIPS2023Dettmers}.

However, while LLMs excel in scalability and efficiency, their strengths can be enhanced by integrating the nuanced understanding and subjective judgments that humans provide, particularly in complex or domain-specific contexts \cite{wang2023largelanguagemodelsfair}. 
For example, human evaluators can detect subtle inconsistencies or interpret ambiguous explanations in tasks that require deeper contextual understanding, which an automated method may overlook \cite{MILLER20191,hoffman2019metricsexplainableaichallenges}.

To leverage the strengths of both approaches, we propose a hybrid evaluation method that combines the scalability of LLM-based evaluations with the depth of human assessment. In this approach, a strong LLM such as GPT-4 is used to evaluate a broad set of model outputs, ensuring cost-effective and consistent initial assessment. Human evaluators then validated a representative subset of these outputs to uncover overlooked biases or misalignments. This approach integrates the efficiency and scalability of automated methods with the reliability of human judgment, particularly in nuanced cases.

By adopting this hybrid methodology, we balance efficiency and accuracy, ensuring that the evaluation process captures both objective metrics and subjective preferences effectively. This dual-layered evaluation enables comprehensive assessments while addressing the inherent limitations of relying exclusively on either LLMs or humans.

\subsubsection{LLM-as-a-judge Evaluation}\label{sec:categoriesEvaluationQuestions}

\begin{figure}
	\centering
	\includegraphics[height=6.4cm]{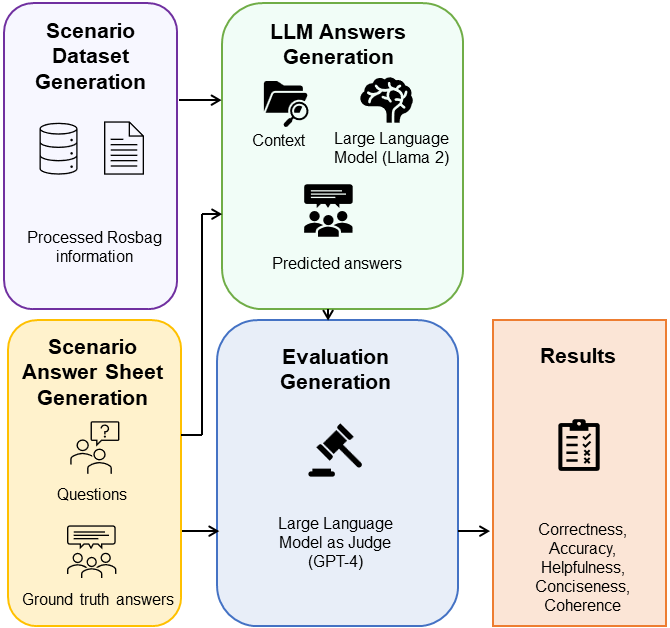}
	\caption{Explainability evaluation.}
	\label{fig:llm_evaluation}
\end{figure}

The need for a robust LLM aligned with human preferences led us to select the GPT-4 model to evaluate the answers produced by our explainability component. Through LLM-as-a-judge, we can obtain not only scores but also explanations, making their outputs interpretable. The entire process is illustrated in Figure \ref{fig:llm_evaluation}.

LLMs exhibit a certain degree of indeterminism, defined as a model's ability to offer different answers when asked the same question or prompt several times. This behavior is derived from their training, in which the models learn to generate outputs by sampling from probabilistic distributions over possible word sequences, thereby introducing randomness to their responses. The effect of indeterminism must be considered in LLM evaluations.

In addition, the use of RAG for question-answering systems is one of the most difficult LLM use cases for evaluation. A basic RAG setup includes two main sources of error: the retriever and the LLM. The evaluation of the retriever aims to determine the relevance of document selection for a query. To this end, we conducted an experimental study with 16 different questions on the data generated and processed during the navigation task in the three previously described scenarios. This set of questions has been evaluated six times for each scenario, giving a total of 288 questions and answers to perform the analysis. This approach ensures a thorough examination of the performance of the RAG system in various scenarios, demonstrating the potential strengths and weaknesses of our solution.

Using the context provided by the Rosbag data and a given evaluation question, our approach includes an evaluation component that assesses the generated explanations by comparing the prediction made by the LLM against an annotated ground truth. The comparison yields a correctness score of 1 if the prediction matches the ground truth, and 0 otherwise. This binary metric reflects the extent to which retrieved information supports accurate explanations. The final score, computed as the average across the dataset, measures the retriever’s overall effectiveness in providing context that leads to correct, grounded responses.

A comprehensive evaluation of an LLM should consider multiple metrics including correctness and other aspects. A language model should produce text that is not only correct but also logically structured and easy to understand. The responses must be appropriately detailed without unnecessary verbosity. Furthermore, the LLM must not only provide correct information but also deliver it in a meaningful and valuable manner to the user. 

\begin{figure*}
	\centering
	\includegraphics[height=15.5cm]{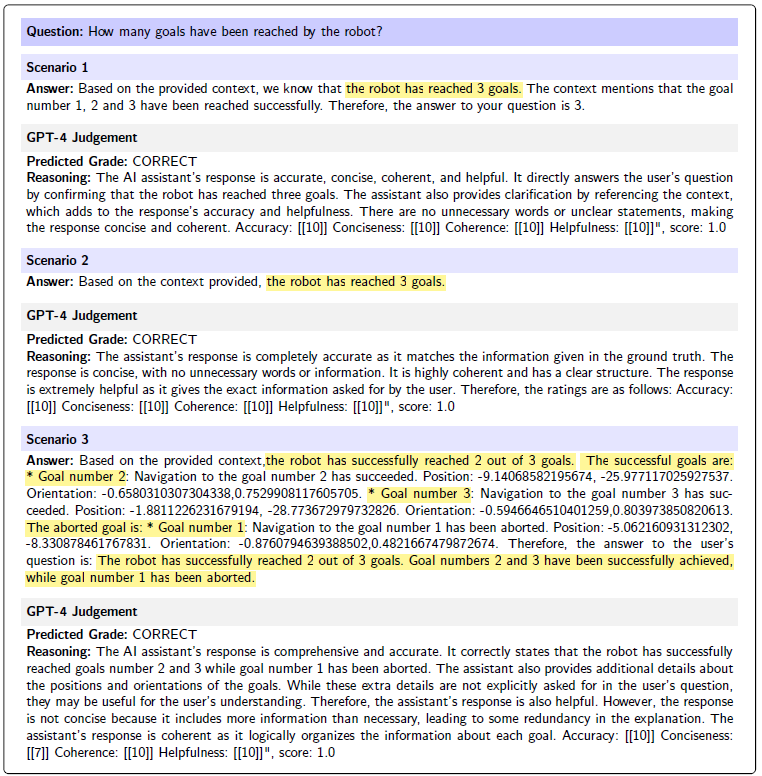}
	\caption{Examples of correct answers, RAG, and criteria-based evaluations obtained across the three scenarios for the question 'How many goals have been reached by the robot?' with LLM-as-a-judge evaluation.}
	\label{fig:QA_Example}
\end{figure*}

Therefore, to assess the performance of the selected LLM on specific attributes, this approach also includes a criteria-based evaluation of accuracy, conciseness, coherence, and helpfulness. The selected criteria metrics are described in Table \ref{table:criteria-evaluation}. This structured assessment offers a means to assess the model's effectiveness in generating responses, providing a broader understanding of the model's capabilities and its alignment with the scenarios used. By employing a scorecard defined for each of these features, our solution assigns scores between 0 and 10 to the responses it generates. The use of this scorecard facilitates a detailed and quantitative analysis of LLM output, ensuring a comprehensive understanding of both its strengths and areas for improvement. A complete example of the answers and evaluations obtained for a selected question in the three scenarios is shown in Figure \ref{fig:QA_Example}.

\begin{table*}[ht!]
    \caption{Criteria-based evaluation metrics in LLM-as-a-judge evaluation.}
    \label{tab:answers}
    \begin{center}
            \begin{tabular}{ p{2.8cm}  p{13.8cm}}
                \toprule
                Metric & Description \\
                \midrule
Accuracy & Measures how well the response aligns with the reference, ranging from completely inaccurate to fully accurate.\\
Helpfulness & Assesses the informational value of the response, from uninformative to highly insightful and comprehensive.\\
Conciseness & Evaluates the brevity and relevance of the response, from overly verbose to succinct and focused.\\
Coherence & Judges the logical structure and flow, from disorganized or inconsistent to well-structured and logically connected.\\
                \bottomrule
            \end{tabular}
    \end{center}
    \label{table:criteria-evaluation} 
\end{table*}

\subsubsection{Human Evaluation}\label{sec:HumanEvaluation}
To validate the explanation results generated by the GPT-4 model, we conducted an online study with 17 participants with diverse demographics. The age distribution ranged from 18 to 55 years, with the majority (47.06\%) in the 26-35 age group, followed by 35.29\% in the 18-25 age group, 11.76\% in the 36-45 age group, and 5.88\% in the 46-55 age group.
In terms of gender, 62.5\% were male and 37.5\% were female. The participants' highest level of education varied, with 37.5\% holding a PhD, 37.5\% having a master's degree, and 25\% possessing a bachelor's degree. Regarding expertise in explainability of autonomous agents, 37.5\% were beginners, 31.25\% had intermediate expertise, 12.5\% were advanced, and 18.75\% had no experience.

The evaluation process involved three questionnaires, with one corresponding to each scenario. The participants assessed 16 explanations per scenario, resulting in a total of 48 explanations evaluated by each individual. Each explanation was rated using five questions derived from the metrics outlined in Table \ref{table:QuestionMetricsHumanEval}, based on the Explanation Satisfaction Scale proposed by Hoffman et al. \cite{hoffman2019metricsexplainableaichallenges}. This scale is a widely recognized framework for measuring human responses to XAI systems. To ensure that the evaluation aligned with human preferences, we used a 5-point Likert scale. Answers in this approach are distributed in 1 (I disagree strongly), 2 (I disagree somewhat), 3 (I'm neutral about it), 4 (I agree somewhat), and 5 (I agree strongly).

In addition to the explanation evaluations and optional demographic information, the questionnaire provided general instructions describing the robot's task and included a map for each scenario. The questionnaires were distributed via a mail list, and participants were informed about the anonymity and research purposes of the study. Consent was obtained upon submission of their answers. Participation was voluntary, which made the process cost-effective but also introduced potential drawbacks, such as longer response times and biases that may arise from factors such as age, education level, or familiarity with autonomous system explainability approaches.

Statistical analysis is essential in human evaluations to ensure rigor and reliability, particularly when interpreting subjective judgments and addressing variability among participants. Likert-scale data, which is ordinal, requires careful statistical treatment to identify significant trends and differences while avoiding misinterpretation of the results. By applying appropriate statistical methods, we strengthen the credibility of our conclusions and provide a robust evaluation of LLM’s results across scenarios.


A series of statistical tests was applied to ensure the reliability and robustness of the evaluation. Reliability analysis of the scale criteria was conducted using Cronbach's Alpha, a robust metric for assessing internal consistency and ensuring that the scale reliably measures the intended constructs, with a commonly accepted threshold for reliability of \( \alpha > 0.7 \) \cite{ujang2018review}. 

To further explore the relationships among variables in the dataset, correlation analysis was performed using Pearson correlation matrices,  which provide insights into the strength and direction of the linear relationships between variables \cite{Benesty2009}.

For inferential analysis, a normality check was conducted using the Shapiro-Wilk test to determine whether the data met the assumptions required for parametric tests. As the data violated these assumptions, the non-parametric Friedman test was employed \cite{razali2011power}. The Friedman test is particularly suited for analyzing repeated measures on the same participants, as it does not require data to follow a normal distribution. This test assessed whether significant differences existed in participant responses across the three scenarios, offering a reliable framework for understanding the impact of different conditions on the perceived quality and effectiveness of explanations.

Finally, a post-hoc analysis using the Nemenyi test was performed to examine participant behavior in specific explanations where significant differences were observed, offering a deeper understanding of the underlying patterns in their responses. Statistical analyses were performed through Python scripts (version 3.10.12), utilizing libraries such as pingouin (version 0.5.5) for Cronbach's Alpha, scipy.stats (version 1.14.1) for the Shapiro-Wilk and Friedman tests, and pandas (version 2.3.3) for data manipulation. This approach ensured accuracy and reproducibility during the text execution.

\begin{table*}[ht!]
\centering
\caption{Questions and metrics used for human evaluation.}
\begin{tabular}{ll}
\toprule
 Metric & Question  \\
\midrule
Understandability & From the explanation, I understand how the task proceeded.\\
Satisfaction & The explanation is satisfactory.\\
Informativeness & The explanation has sufficient detail. \\
Completeness & The explanation seems complete. \\
Usefulness & The explanation is useful to my goals.\\
\bottomrule
\end{tabular}
\label{table:QuestionMetricsHumanEval}
\end{table*}

\subsubsection{Categories and Evaluation Questions}

To evaluate our explainability proposal in the context of ROS 2 message interpretation, we defined a set of questions covering diverse aspects of the navigation process over the \textit{how}, the \textit{what}, the \textit{which}, the \textit{when} and the \textit{where} dimensions. These questions provide insights into the different stages and components of the three scenarios. To structure and improve the understanding and analysis of this set of questions, we organized them into five categories. These categories range from high-level overviews of navigation to detailed inquiries regarding trajectory planning, goal completion, and BT functioning. These categories also delve into the causes of unexpected behavior or errors in navigation, improving the explainability of the system whenever an unexpected action occurs. Through these questions, we aim to address key concerns in XAI, including understanding why decisions are made, establishing causality between actions, determining success or failure, identifying timing, and disguising the methods by which achievements are realized \cite{gunning2019darpa}.

The set of categories, their descriptions and questions included in each one are described next.
\paragraph{Navigation Process Overview.} These questions provide an overview of the navigation process captured at the Rosbag file. Understanding what has occurred and how the navigation task has proceeded provides a high-level context that sets the stage for more detailed questions in later categories. This category comprises the following questions:
\begin{enumerate}
    \item What has happened in this ROS 2 log regarding navigation?
    \item How has the navigation task proceeded?
\end{enumerate}
\paragraph{Trajectory Planning and Re-planning.} These questions delve into the specifics of trajectory planning, which is a critical aspect of navigation. They address whether the robot has re-planned its trajectory, the reasons behind this behavior, and whether it has encountered any obstacles. These details are necessary to understand how a robot dynamically adjusts its path. The questions included in this category are:
\begin{enumerate}\addtocounter{enumi}{2}
    \item Has the robot re-planned an alternative trajectory during navigation?
    \item Why did the robot re-plan the route?
    \item Did the robot find any obstacles during the navigation?
\end{enumerate}
\paragraph{Goal Completion and Navigation Task Status.} These questions focus on the completion status of the navigation goals. Knowing how many targets have been reached, whether the robot has completed the navigation task, and when the navigation task ended provides insights into the overall progress and success of the robot's task. This category includes the following questions:
\begin{enumerate}\addtocounter{enumi}{5}
    \item How many goals have been reached by the robot?
    \item Has the robot completed the navigation task?
    \item When has the robot ended the navigation task?
    \item Have all objectives been successfully achieved or have any been cancelled or aborted?
\end{enumerate}
\paragraph{Specifics about Goals and Locations.} These questions examine the specifics of the goals and locations. Understanding the location of goals, the robot's linear velocity during navigation, its initial position and orientation, and the velocity after receiving a goal contribute to a detailed understanding of the robot's movements and performance. The questions included in this category are:
\begin{enumerate}\addtocounter{enumi}{9}
    \item Where is the second location or goal located?
    \item What was the linear velocity when navigating to goal pose number 2?
    \item What are the initial position and orientation of the robot?
    \item What was the linear velocity of the robot after receiving goal number 1?
\end{enumerate}
\paragraph{Nav2 Behavior Tree and Node Status.} These questions provide information about the underlying behavior tree and the state of its nodes in the navigation system. Knowing the specific node responsible for determining a viable path, identifying any failed nodes during navigation, and understanding the use of behavior tree nodes, contribute to a deeper understanding of the inner workings of the navigation system. This category includes:
\begin{enumerate}\addtocounter{enumi}{13}
    \item What is Nav2 Behavior Tree's node to determine a viable path from a starting point to a specified target pose or location?
    \item Did any node from the Nav2 Behavior Tree fail during navigation?
    \item Which Behavior Tree nodes were used during navigation?
\end{enumerate}

\section{Experimental Results and Discussion}\label{sec:experimental_results}
This section summarizes the results of the experiments involving both the black box-like engine and explainability component when using our solution in the previously depicted assessment scenarios.

\subsection{Message Loss Rates}
\begin{figure}[hbt!]
    \centering
    \subfigure{\includegraphics[scale=.50]{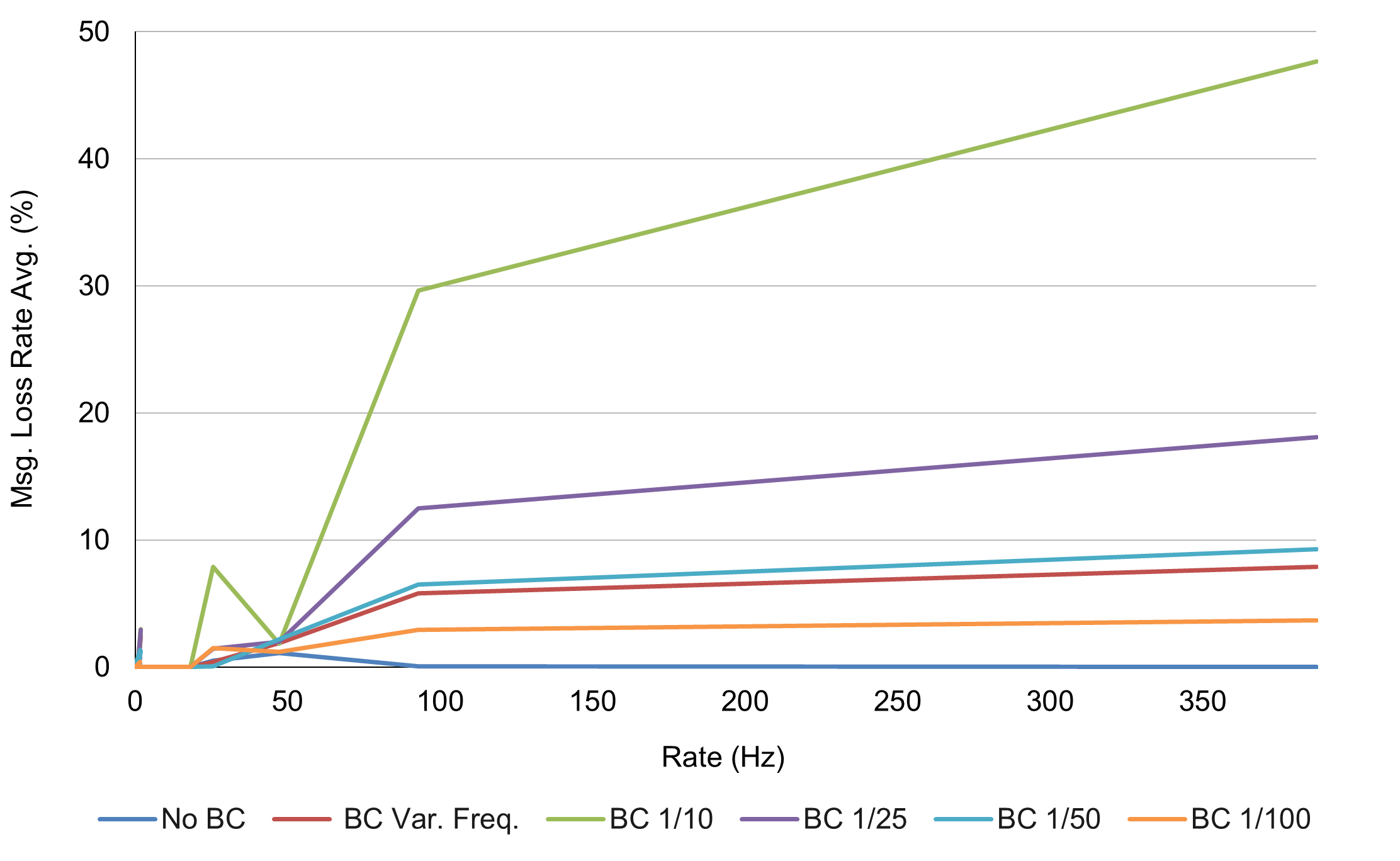}} 
    \subfigure{\includegraphics[scale=.50]{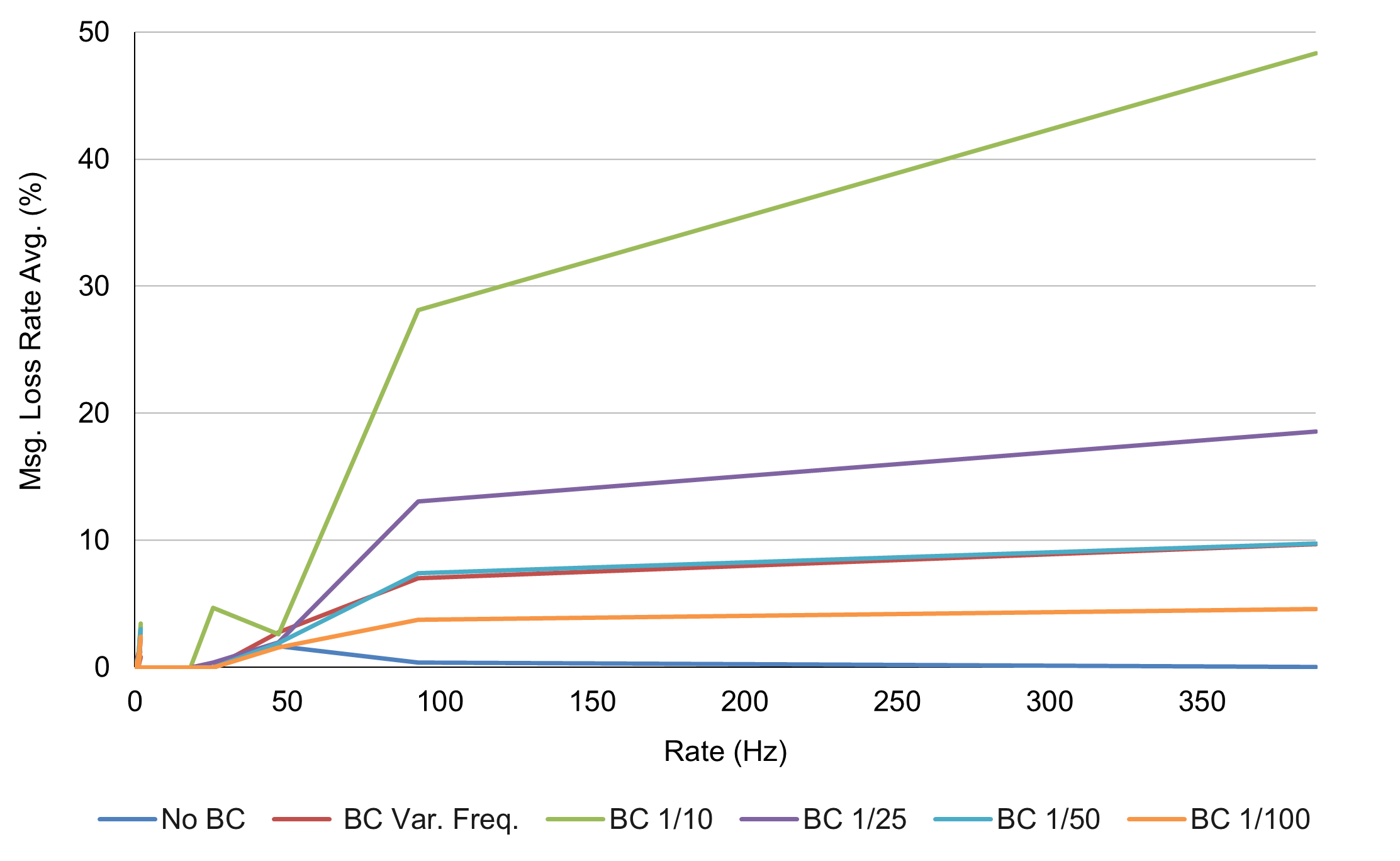}} 
    \subfigure{\includegraphics[scale=.50]{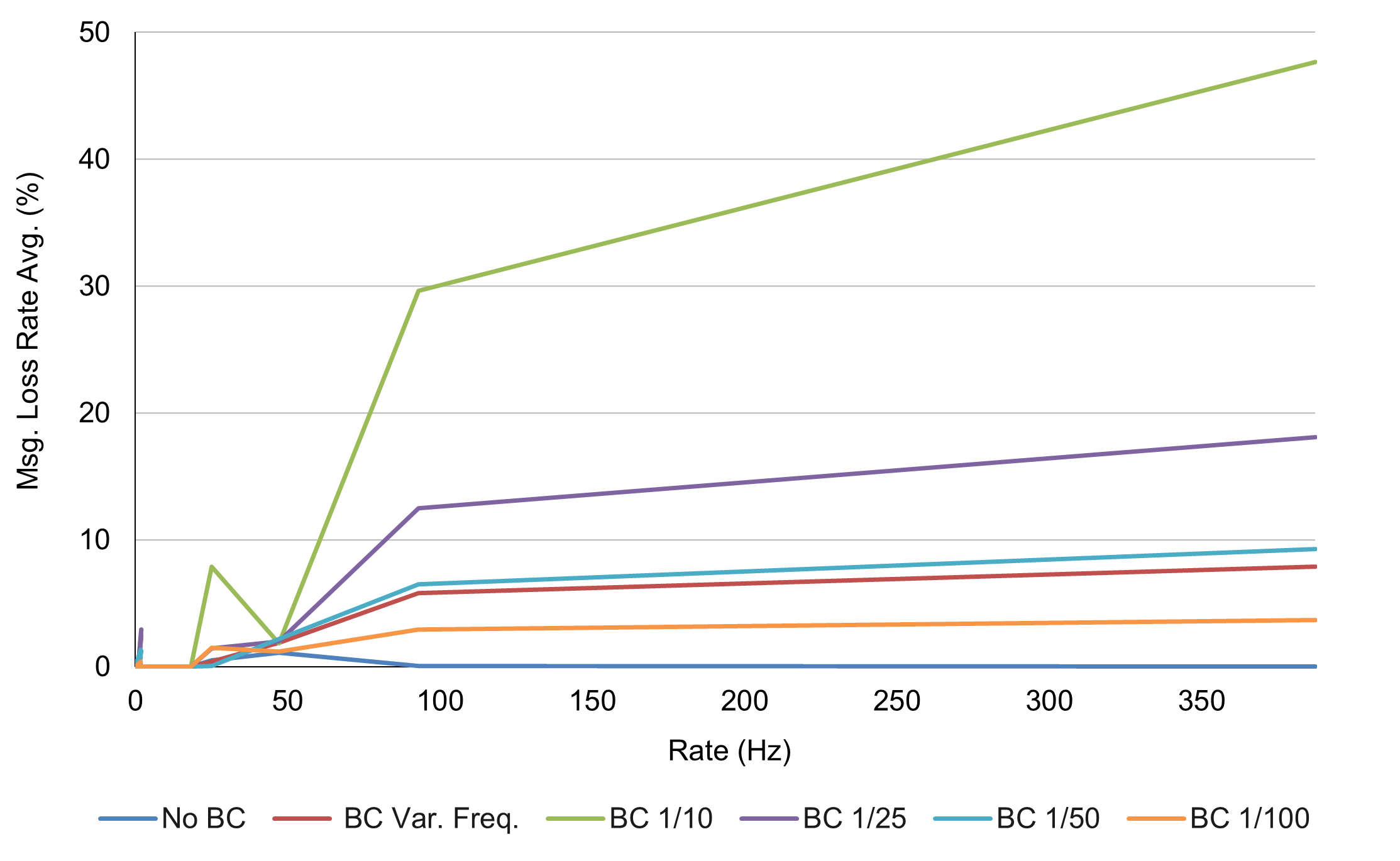}}
    \caption{Message loss rates in Rosbag Files under several anti-tampering techniques. Comparative visualizations for: (a) Scenario 1, (b) Scenario 2, and (c) Scenario 3.}
    \label{fig:msgs_loss_rates}
\end{figure}

Figure \ref{fig:msgs_loss_rates} shows the percentage of lost messages in the Rosbag file for the experiments detailed in Section \ref{sec:evaluation}, capturing the topics outlined in Table \ref{table:ROS2Topics}, attending to their publication rates, and to the approaches selected to store integrity proofs in the blockchain. The results across the three scenarios demonstrate a consistent trend, showing uniform message loss patterns in the creation of immutable Rosbag files.

Choosing to store a single message hash for every 100 messages results in a loss of less than 5\% compared with avoiding anti-tampering measures in the Rosbag recording process, even for topics with rates exceeding 300 Hz. This approach is the most efficient option for integrating anti-tampering techniques into our accountability solution.

Alternative strategies, such as saving one proof every 50 messages or adjusting proof intervals based on topic rates, achieve losses of less than 10\% for each topic. This is particularly noteworthy, as it ensures the integrity of messages based on their significance in a given scenario or task.

Introducing an integrity proof every 25 messages leads to an overall reduction in message recording, particularly when topic rates surpass 100 Hz. Storing one hash every 10 messages could lead to losses exceeding 30\% for topics with rates higher than or around 100 Hz. Beyond the consistent linear trend of message loss with high topic rates, a slight deviation is observed in this approach when topics fall below 50 Hz. This variation signifies the influence on message recording when computing hash functions for messages nearing 1 MB in size, as observed in messages from the /camera/image\_raw topic, given that the computational complexity of common hash functions, such as SHA-256, increases with input size.

\subsection{System Benchmarking}
\begin{figure}[hbt!]
    \centering
    \subfigure{\includegraphics[scale=.50]{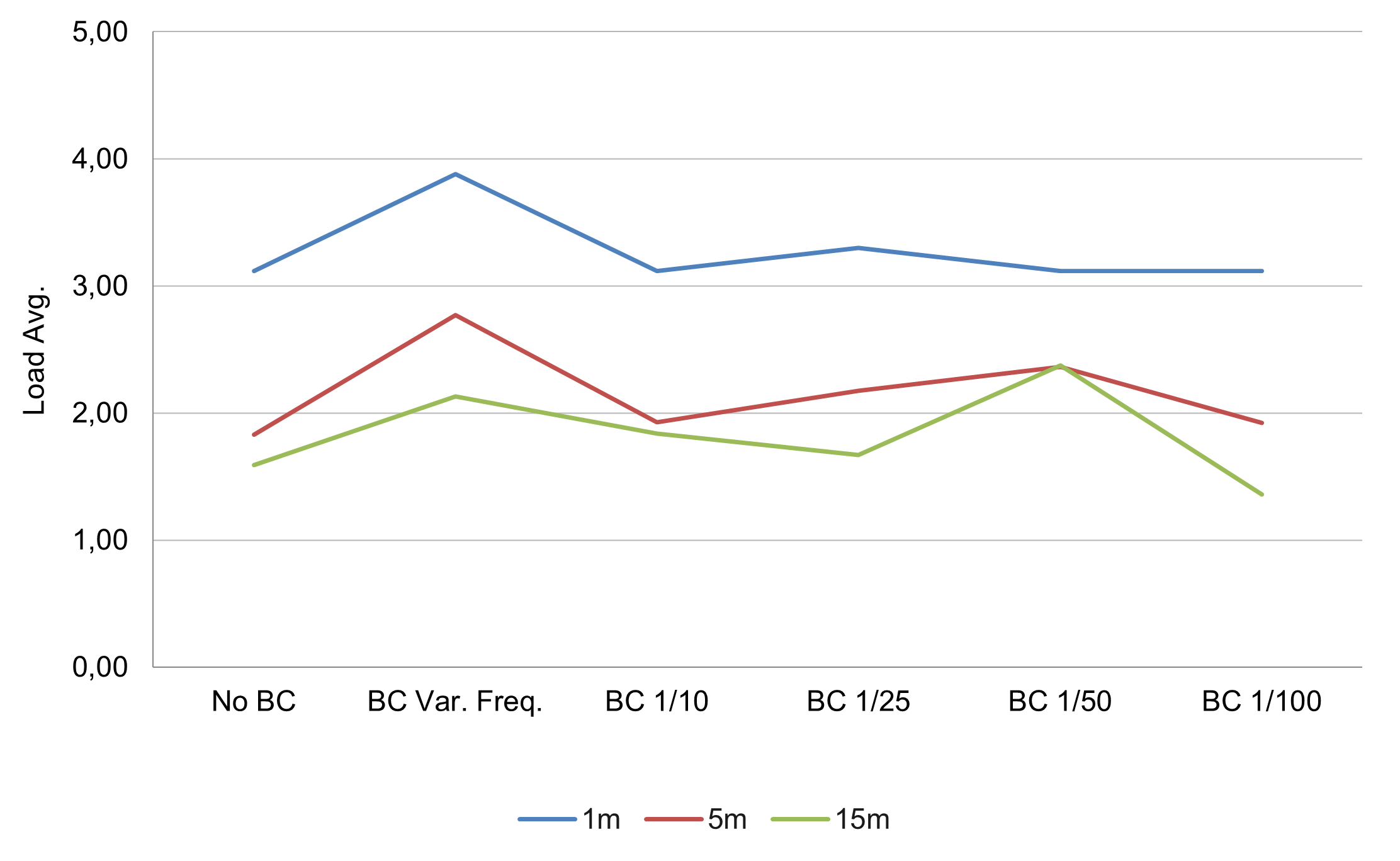}} 
    \subfigure{\includegraphics[scale=.50]{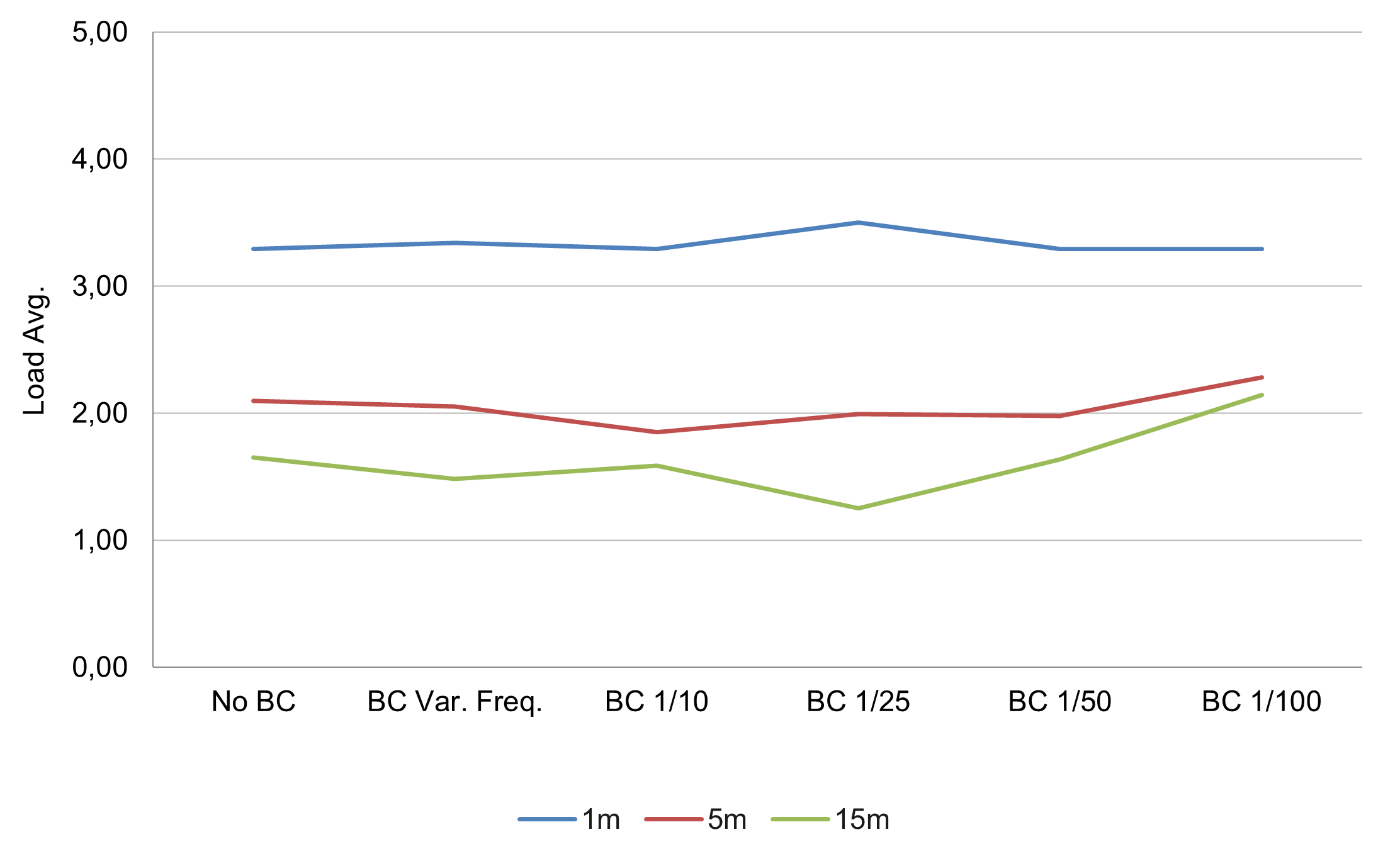}} 
    \subfigure{\includegraphics[scale=.50]{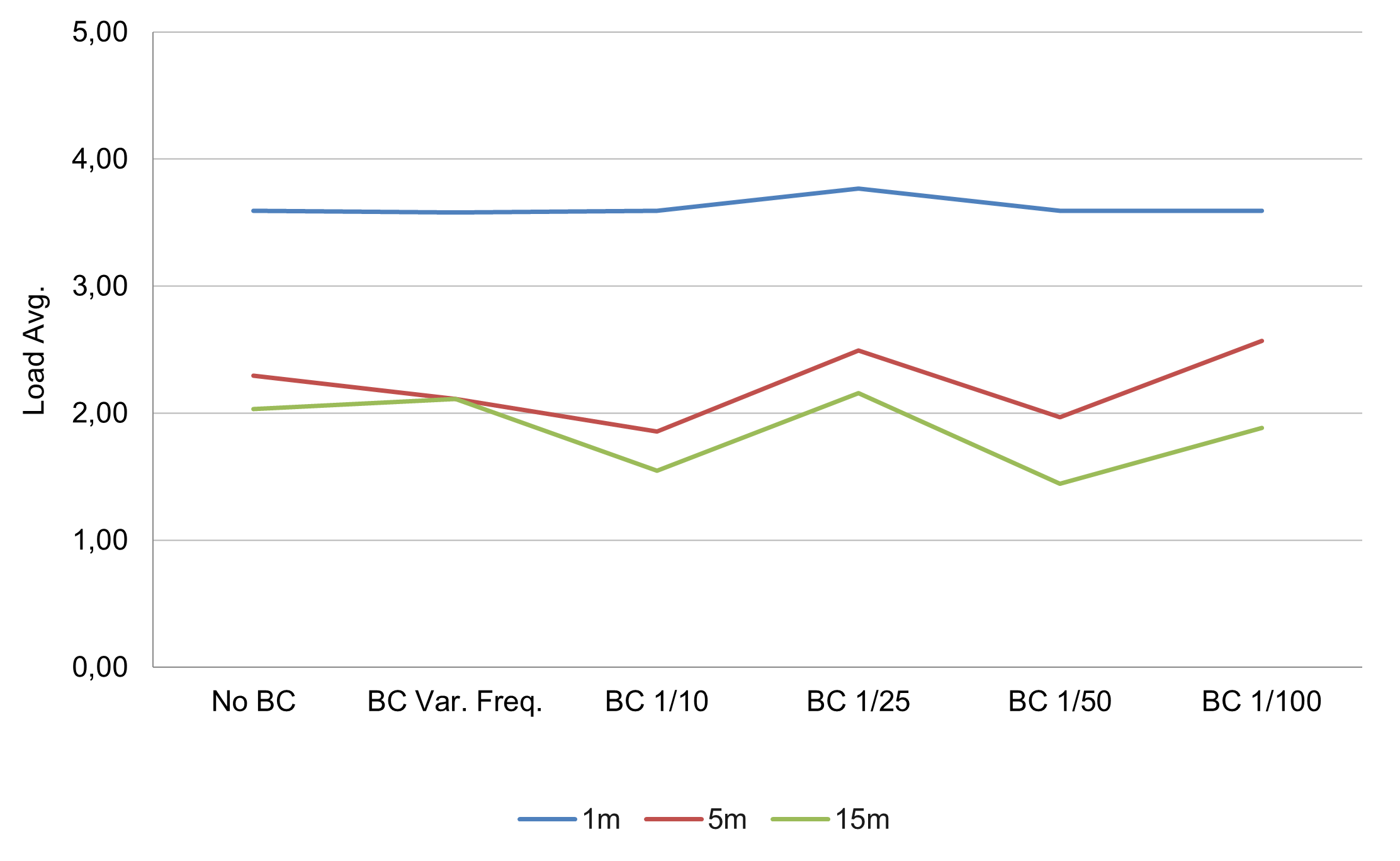}}
    \caption{Load average under several anti-tampering techniques. Comparative visualizations for: (a) Scenario 1, (b) Scenario 2, and (c) Scenario 3.}
    \label{fig:load_avg}
\end{figure}

The system load or, running thread demand on the system as the average number of running and idle threads, is shown in Figure \ref{fig:load_avg}. The average of 1 minute is higher than the average of 5 or 15 minutes in all scenarios, showing a load increase. However, the values of this metric do not exceed the CPU count, which is 6 in our experimental setup, indicating no performance issues or resource saturation. The inclusion of runnable and uninterruptible tasks in this metric means that the load average may increase due to a disk I/O workload, not just CPU demand, giving a more general view of the system overhead.
Other more specific metrics are described next to clarify the possible existence of a bottleneck in the proposed approach.

The CPU consumption by user processes maintains a consistent linear trend across all approaches and scenarios, remaining around 30\%. This level of consumption is generally considered acceptable for resource-intensive processes such as the recording of a Rosbag file, providing a substantial margin in the event of introducing other critical tasks into the system.

Concerning memory consumption, all approaches within the three scenarios exhibit comparable values, differing by no more than 0.5 GB. Consequently, the inclusion of blockchain approaches into our black box recorder does not yield a remarkable impact on RAM consumption in comparison to avoiding the use of anti-tampering techniques in this recording process.

Regarding network traffic, approaches with higher process storage in the blockchain increase this metric, but do not exceed 0,20 MB/s, which is a very affordable quantity. Furthermore, approaches such as storing one hash every 100 messages nearly match the avoidance of storing any proof into the blockchain, indicating that including this approach does not have a remarkable impact on system performance.

Disk write operations do not form a bottleneck in system performance, with all approaches differing by no more than 0,40 MB/s, which is not a very significant impact. In addition, all approaches that perform integrity-proof storage on the blockchain achieve lower or similar results to the approach that avoids anti-tampering operations.

\begin{figure*}[ht!]
    \centering
    \subfigure{\includegraphics[scale=.50]{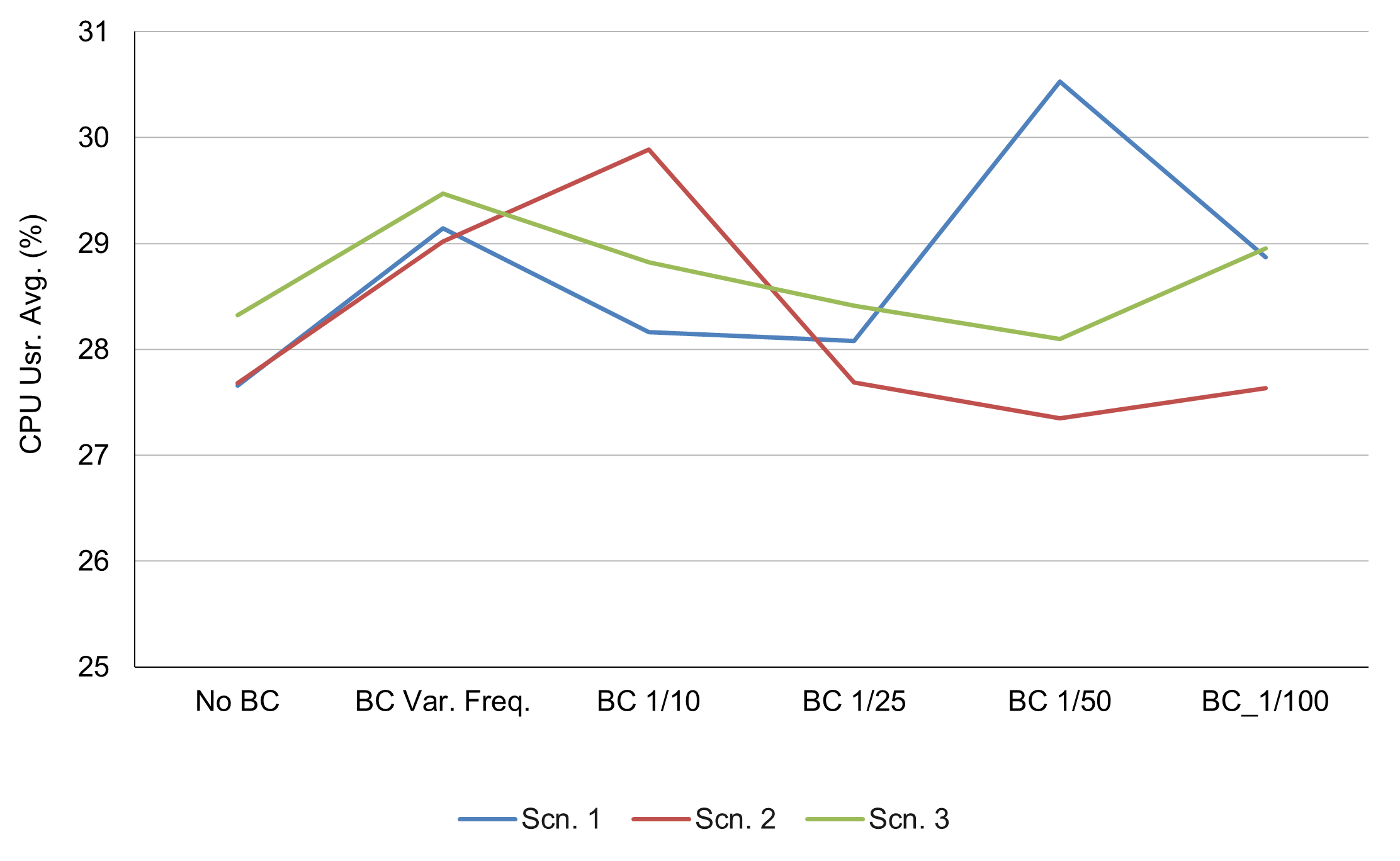}} 
    \subfigure{\includegraphics[scale=.50]{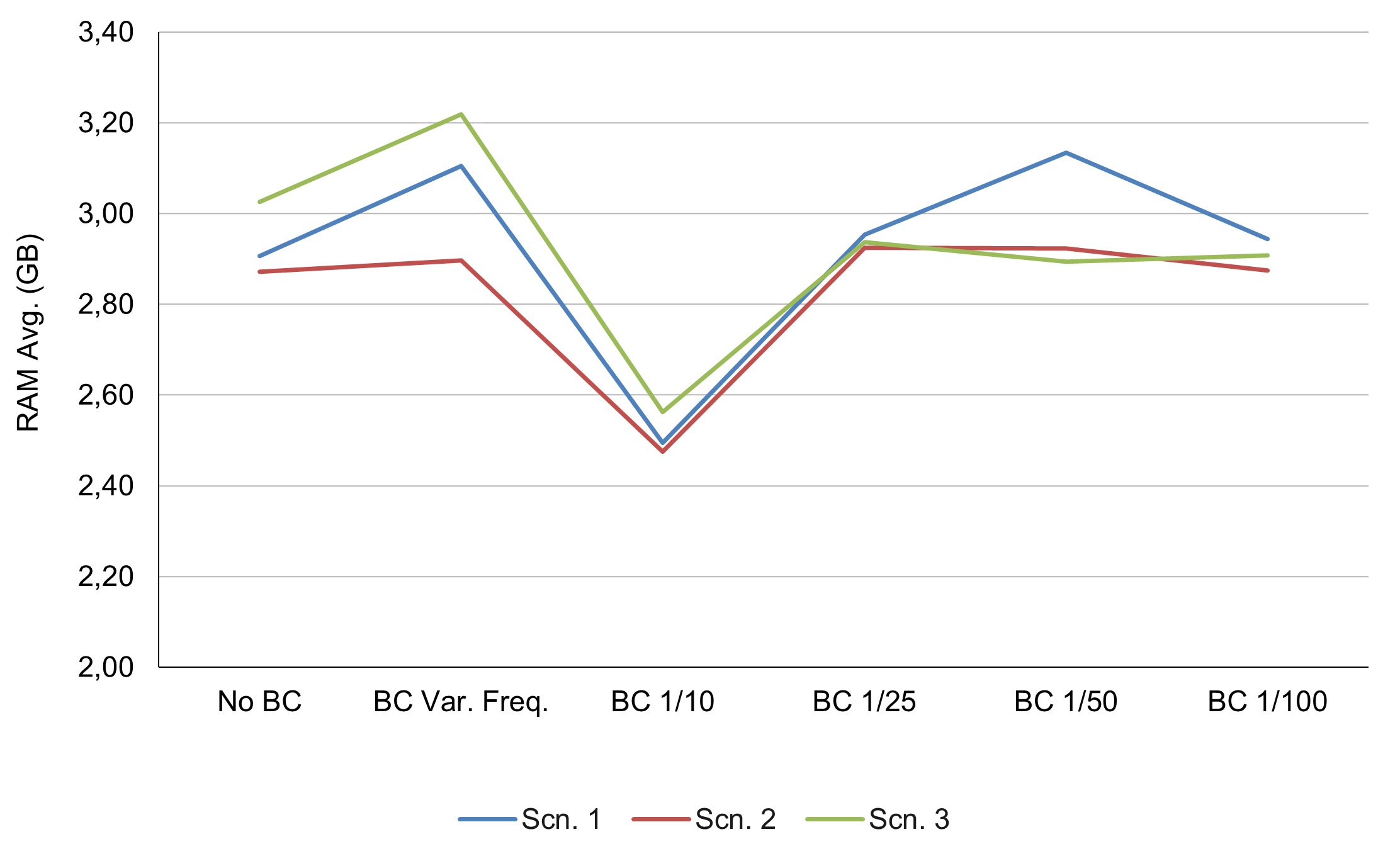}} 
    \subfigure{\includegraphics[scale=.50]{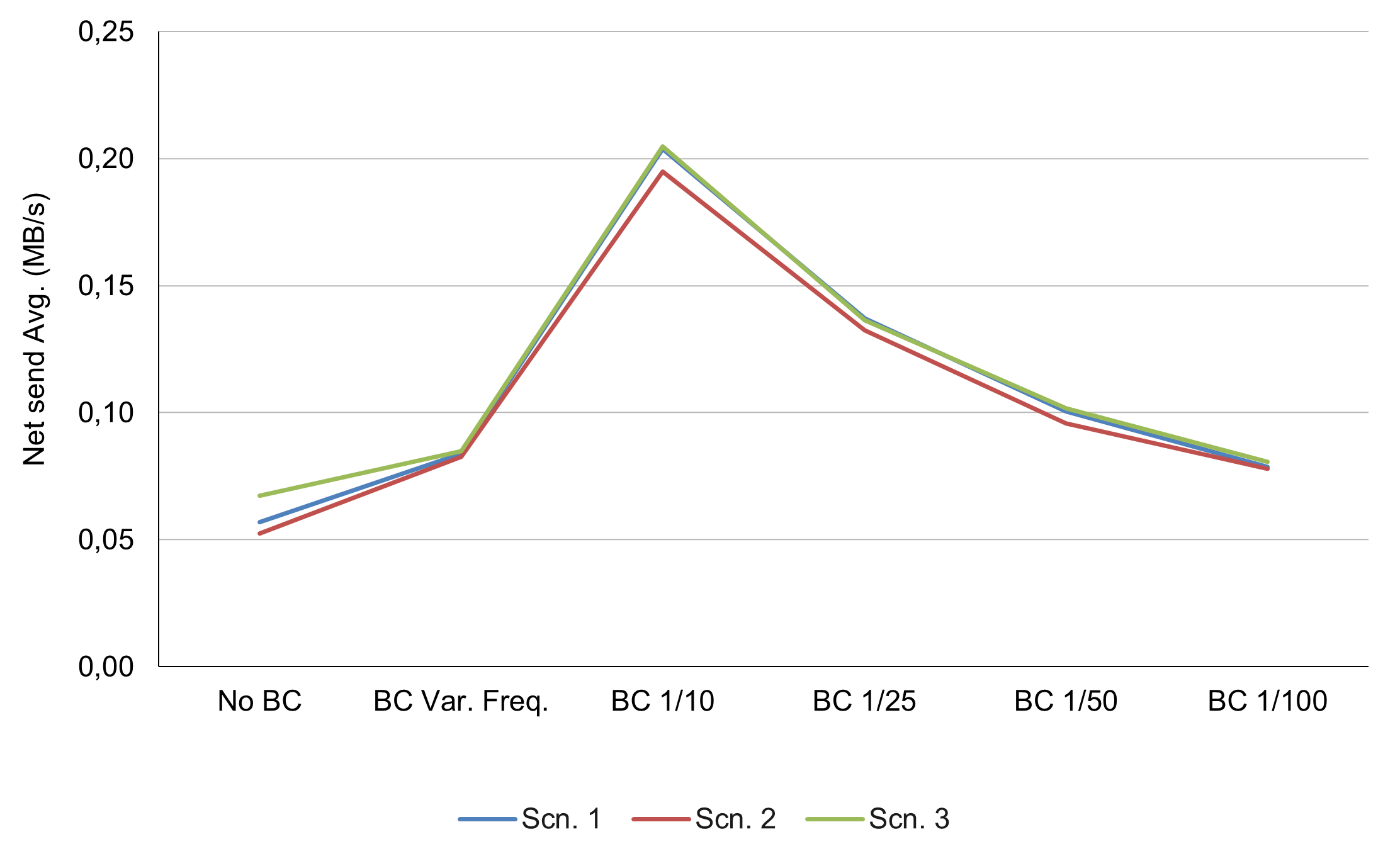}}
    \subfigure{\includegraphics[scale=.50]{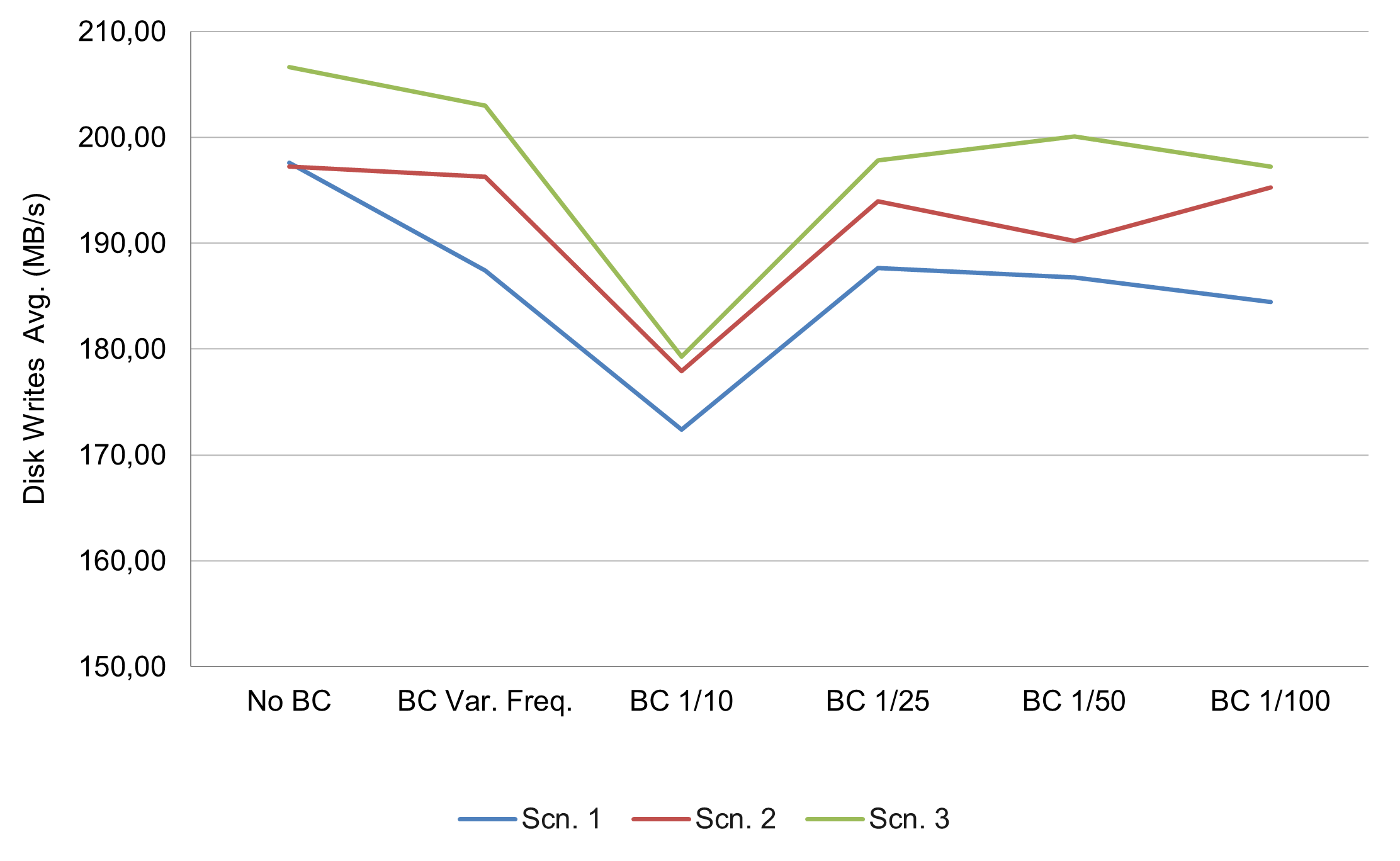}}
    \caption{System benchmarking results: (a) CPU usage of user processes. (b) RAM usage. (c) Sent network traffic. (d) Disk writes traffic.}
    \label{fig:accountability_performance}
\end{figure*}

Figure \ref{fig:accountability_performance} shows the results described.

These findings may encourage the adoption of our approach to ensure integrity in messages from autonomous agents.


\subsection{Correctness Evaluation Using LLM-as-a-judge} \label{correctnessLLMEval}
\begin{figure}
    \centering
    \subfigure{\includegraphics[scale=.50]{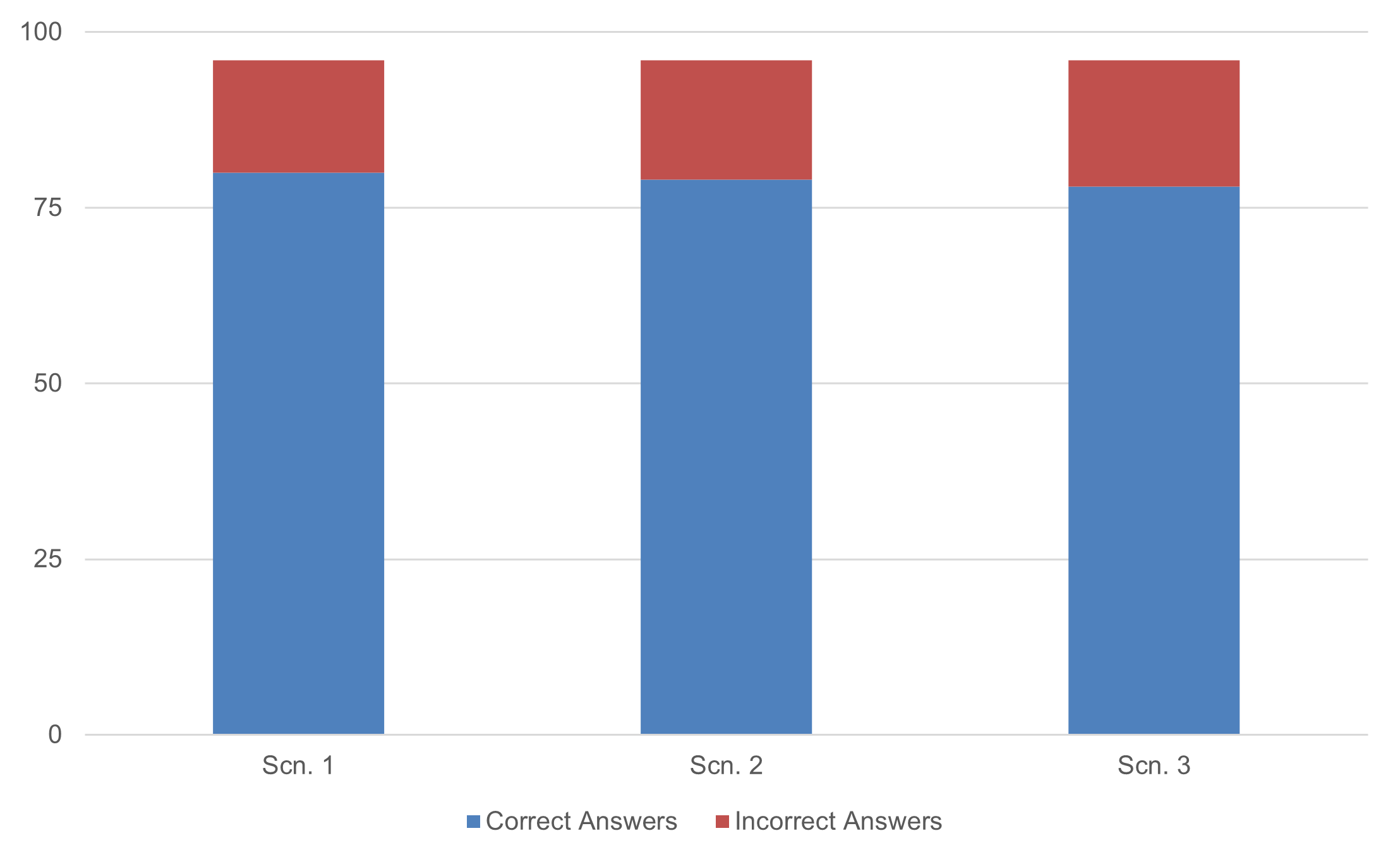}} 
    \subfigure{\includegraphics[scale=.50]{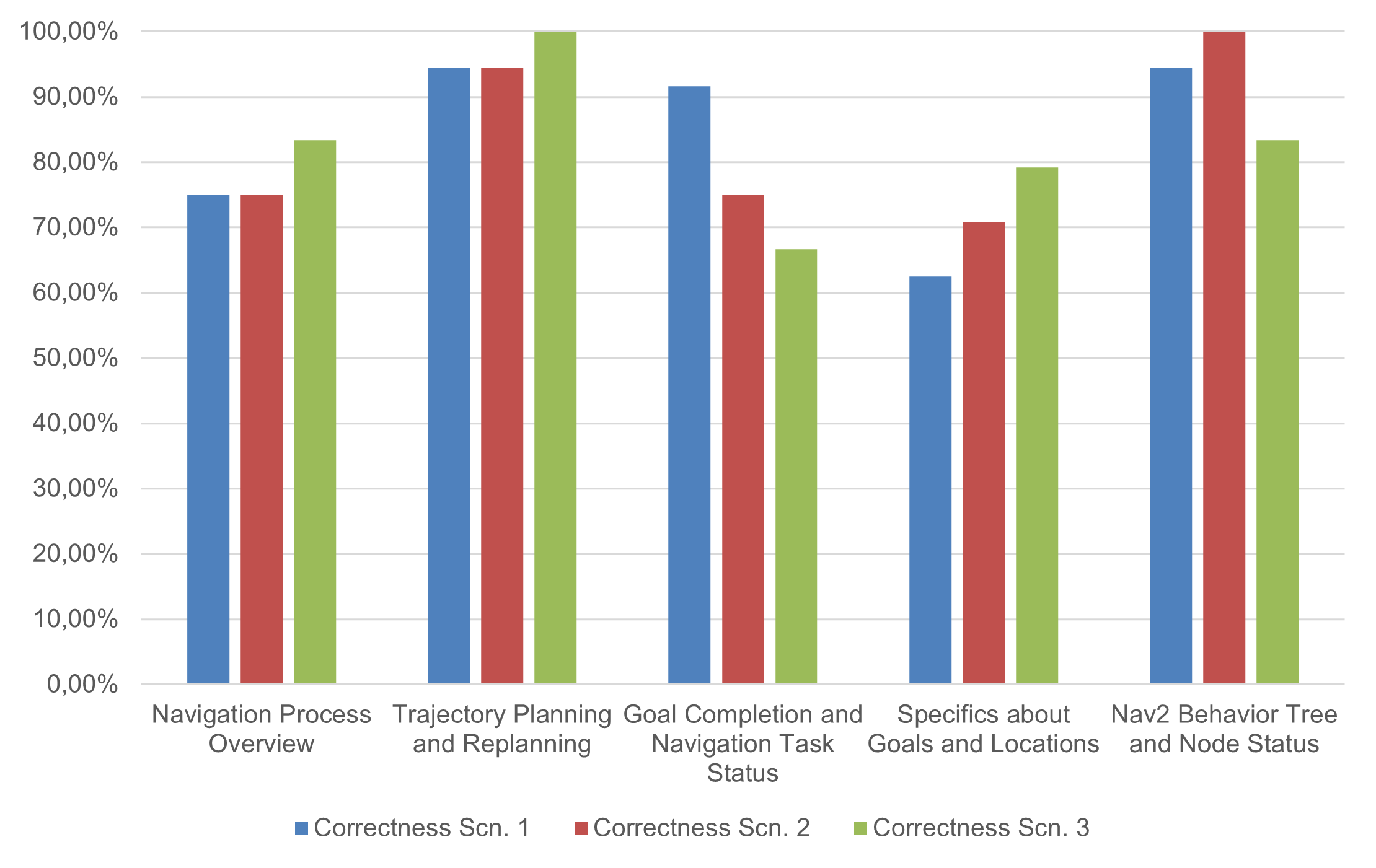}} 
   \caption{(a) Overall correctness in the three scenarios. (b) Correctness by category in the three scenarios.}
   \label{fig:correctness}
\end{figure}

Figure \ref{fig:correctness} shows the results of the correctness of the RAG evaluation, presenting both the overall results and the assessments based on predefined categories. The results demonstrate consistency across the three scenarios, obtaining a significant number of correct answers, even though there has not been prior tuning in the ROS 2 messages for autonomous agents' navigation tasks. The slight percentage of incorrect answers remains within an acceptable threshold, given the specific context employed and the inclusion of general questions.

Regarding individual categories, the correctness in the category \lq Navigation Process Overview' is relatively consistent across all scenarios, indicating a good understanding of events and progress related to navigation. However, the broad nature of the question "What has happened in this ROS 2 log regarding navigation?" initially yielded a correctness rate of less than 65\% in the third scenario. In this instance, the cancellation of a goal resulted in the execution and failure of Nav2 Behavior Tree nodes, which were not included in the other two scenarios and focused on navigation recovery. Therefore, the incorporation in a dataset of information related to the cancellation of a goal results in LLM answers that usually diverge from the ground truth provided, as the model occasionally prioritized BTs information, other times focused on locations, or timestamps. This variability made it difficult to align with the ground truth provided, even if the obtained answers could be considered partially accurate. Introducing the prompt extension "Describe the sequence of events regarding navigation goals and include information about each of them." in Question 1 significantly increased the correctness in the first category, achieving results around and exceeding 75\% in all three scenarios. This underscores the effectiveness of prompt engineering when tackling ambiguous or broad queries.

The results also show consistently high correctness across all scenarios concerning questions related to trajectory planning and re-planning. The model can determine whether the modification of a predicted trajectory has been caused by the occurrence of an obstacle or by other circumstance. Scenarios 1 and 2 also exhibit relatively high correctness in questions related to goal completion and task status. However, the model encounters challenges in Scenario 3, occasionally failing to identify whether a navigation that starts correctly concludes with an abortion or cancellation. This issue may be attributed to a misalignment during the retrieval process, specifically in the selection of the best chunk associated with a particular goal associated with numerous embeddings within a single collection from the vector database.

Correctness in identifying specific targets and locations lacks consistency, with results exceeding 50\% in the first scenario, around 70\% in the second and over 75\% in the third. In particular, three scenarios encounter difficulties in identifying the initial position of the robot before starting the navigation, as the model misaligns the queried position with that assumed by the robot when starting the navigation process. This may also be attributed to deficiencies in the retrieval process. In Scenario 1, understanding the sequence of events or order of goals is a challenge for the model, leading to inaccuracies in specifying the second location or goal. Nevertheless, this challenge is not present in Scenario 3 and is absent or reduced in Scenario 2, demonstrating how differences in data distribution can introduce varied responses even when using specific and unambiguous questions.

Through questions included in the \lq Nav2 Behavior Tree and Node Status' category, we assessed the effectiveness of the model's formal linguistic competence compared to its functional linguistic competence, a feature of present LLMs \cite{mahowald2023dissociating}. While the model excels in identifying BT nodes given its description, it faces challenges in pinpointing these components under specific conditions. As a result, the criteria correctness suffers a drop in Scenario 3, although overall results remain above 75\%. In this scenario, the navigation recovery actions caused by the cancellation of the first goal introduced a challenge in the interpretation of the information by the LLM due to the execution of several Nav2 BT nodes not present in the other scenarios. Particularly, the mistakes stem from the model's confusion in determining whether a previously failed Nav2 BT node has recovered or finished its execution, as well as identifying all nodes that were executed during navigation. This can be attributed to the fact that the nuanced behavior of recovery actions and the dynamic execution of BTs during navigation introduce difficulties that the model may not adequately capture from the training data, underscoring the need for additional fine-tuning in such scenario-specific contexts in case the obtained correctness needs to be improved.

\subsection{Criteria-based Evaluation Using LLM-as-a-judge}
\begin{figure}
    \centering
    \subfigure{\includegraphics[scale=.50]{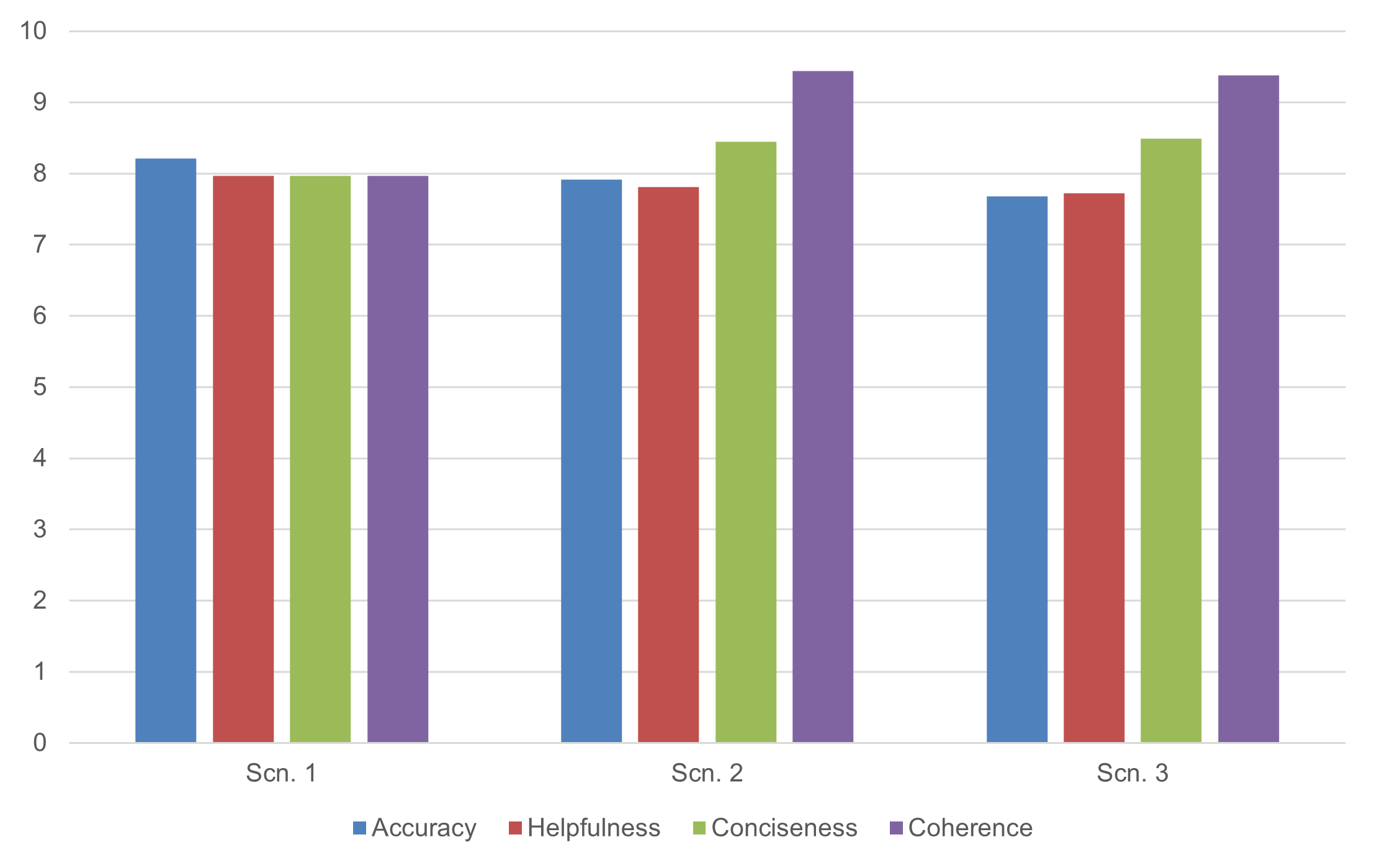}}
   \caption{(a) Overall criteria evaluation in the three scenarios.}
   \label{fig:overall_criteria}
\end{figure}

Figure \ref{fig:overall_criteria} shows the performance of the selected LLM in the evaluated metrics, demonstrating consistency with slight variations in the scores. On average, the answers provided are accurate and in line with the reference, though there might be room for improvement. While responses generally provide useful information, there are instances where additional depth could enhance their utility. In addition, most answers usually include a good balance between conciseness and effectiveness, with clear and logical organization. The overall results show no notable differences in these metrics across the three scenarios. This suggests that messages within the Rosbag related to the introduction of obstacles or cancellation of goals do not notably influence the general criteria evaluation outcomes.

\begin{figure}
    \centering
    \subfigure{\includegraphics[scale=.50]{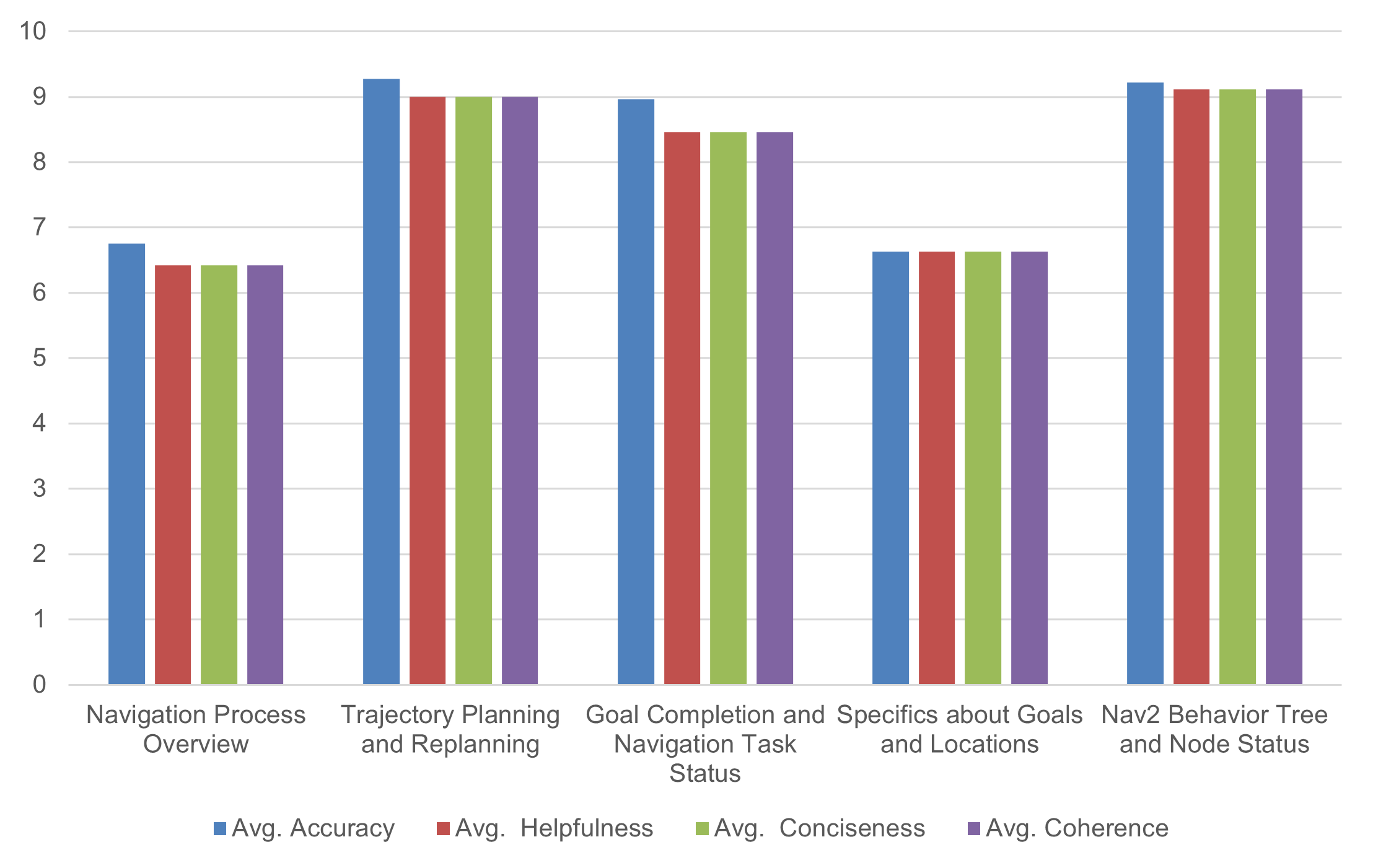}} 
    \subfigure{\includegraphics[scale=.50]{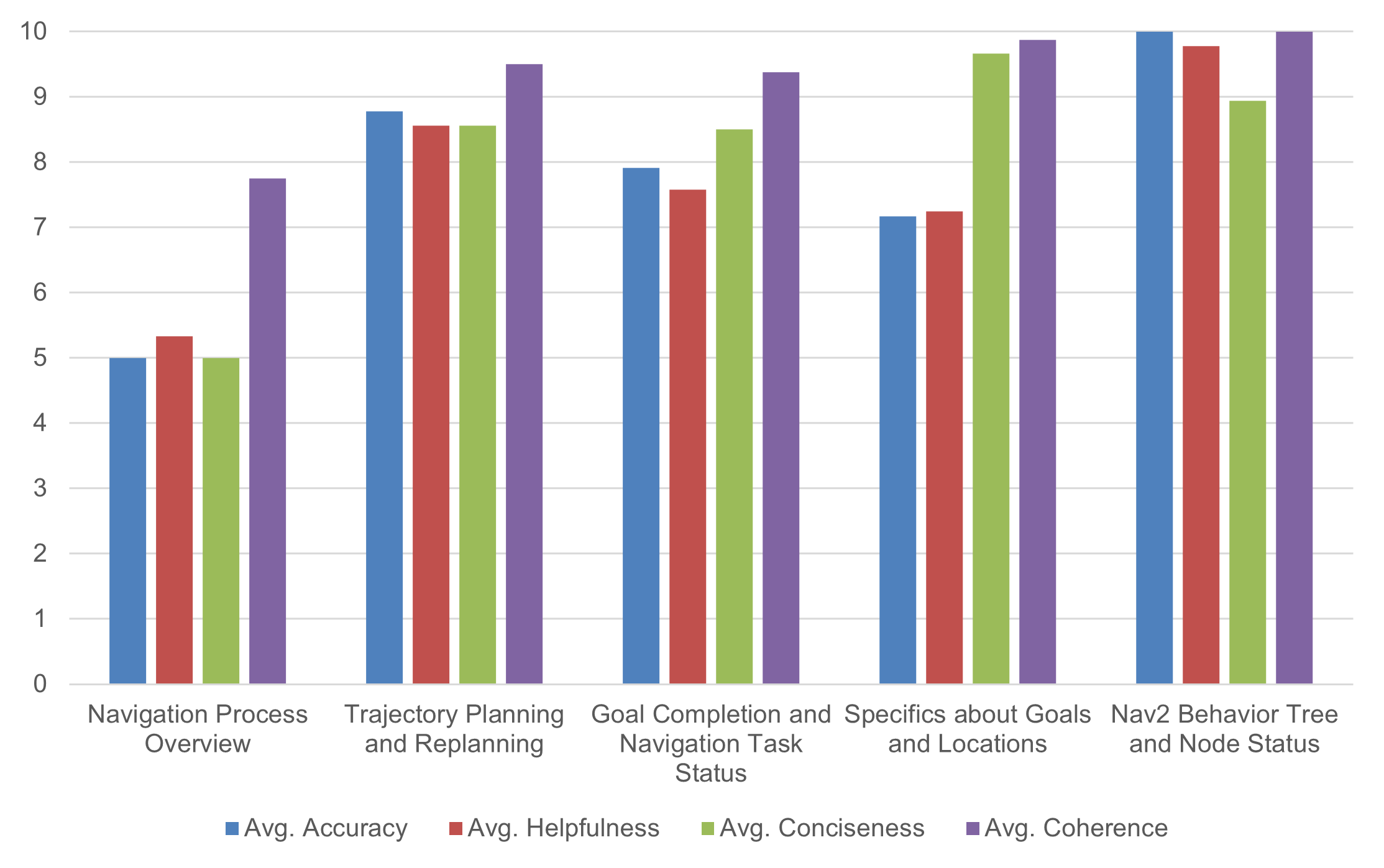}}
    \subfigure{\includegraphics[scale=.50]{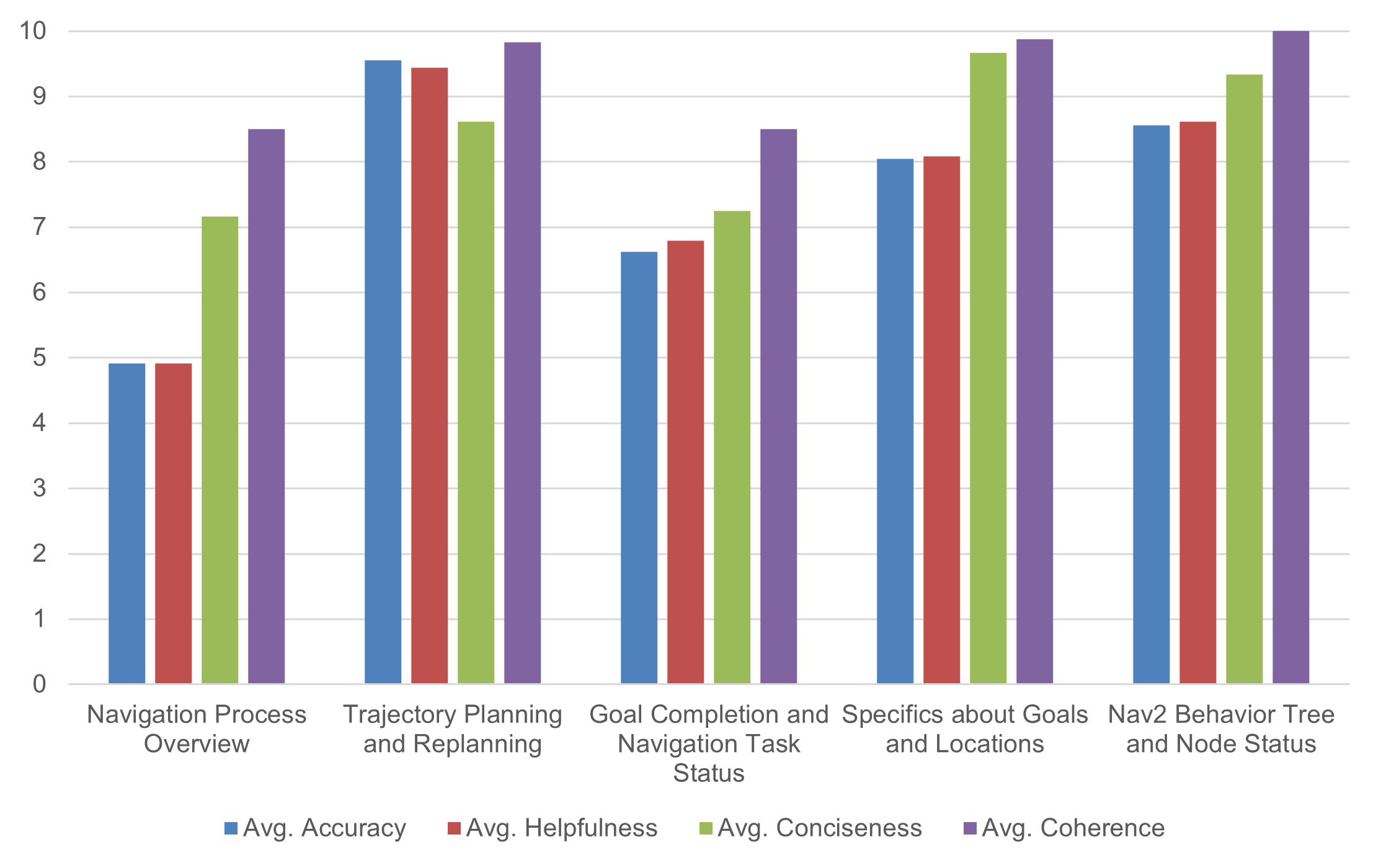}}
   \caption{LLM criteria-based evaluation by categories. Comparative visualizations for: (a) Scenario 1, (b) Scenario 2, and (c) Scenario 3.}
   \label{fig:criteria_evaluation_categories}
\end{figure}

The adaptability of the language model to challenges such as obstacles and goal cancellations regarding the described categories, is shown in Figure \ref{fig:criteria_evaluation_categories}.

Concerning general questions about events and the progression of the navigation task, we observe variations in scores across Scenarios 2 and 3. While coherence and conciseness improve, Scenario 1, which involves direct navigation to the three goals, scores higher in accuracy and helpfulness. This fact suggests that effective communication on general queries benefits from the absence of obstacles or goal cancellations, enhancing overall understanding.

In the category \lq Trajectory Planning and Re-planning', questions are effectively addressed across all three scenarios, showing scores above 8 in every metric, even when obstacles or goal cancellations are introduced. These results highlight the system's capability to manage trajectory-related queries effectively despite these challenges.

Regarding the category \lq Goal Completion and Navigation Task Status' category, Scenarios 1 and 2 effectively convey information about goal completion and navigation task progress. Nevertheless, the slightly lower scores in Scenario 3 indicate challenges in effectively communicating the completion status when a goal is aborted.

In the \lq Specifics about Goals and Locations' category, the system provides valuable goal and location information, even in the presence of obstacles or goal cancellations. This suggests a strong ability to understand and localize specific details regardless of the challenges introduced in each scenario.

The results regarding questions included in \lq Nav2 Behavior Tree and Node Status' category demonstrate effectiveness in communication across all scenarios and metrics.  This emphasizes the potential of BTs to complement language models in explainability, offering insight into decision-making processes during navigation tasks.

In summary, although the results can be improved in accuracy for general questions, the language model shows adaptability across different navigation scenarios, including those with obstacles and task changes. Consistency in conveying information related to Nav2 BTs suggests a solid understanding of underlying navigation processes. The model also performs well when providing information about trajectory planning, goal completion, and specifics about goals and locations, regardless of the specific challenges introduced in each scenario. Nevertheless, although evaluation using LLMs excels in scalability and efficient resource usage, it is still recommended to test a subset of questions with human evaluators. This ensures a deeper understanding of how well the approach aligns with human expectations and communicates effectively, particularly in scenarios involving complex or nuanced details.

\subsection{Human Evaluation of LLM-generated Explanations}
This section presents an analysis of the results of the human evaluation conducted with 17 participants, as stated in Section \ref{sec:HumanEvaluation}.

\subsubsection{Distribution Analysis}
\begin{figure*}
	\centering
	\includegraphics[height=12.0cm]{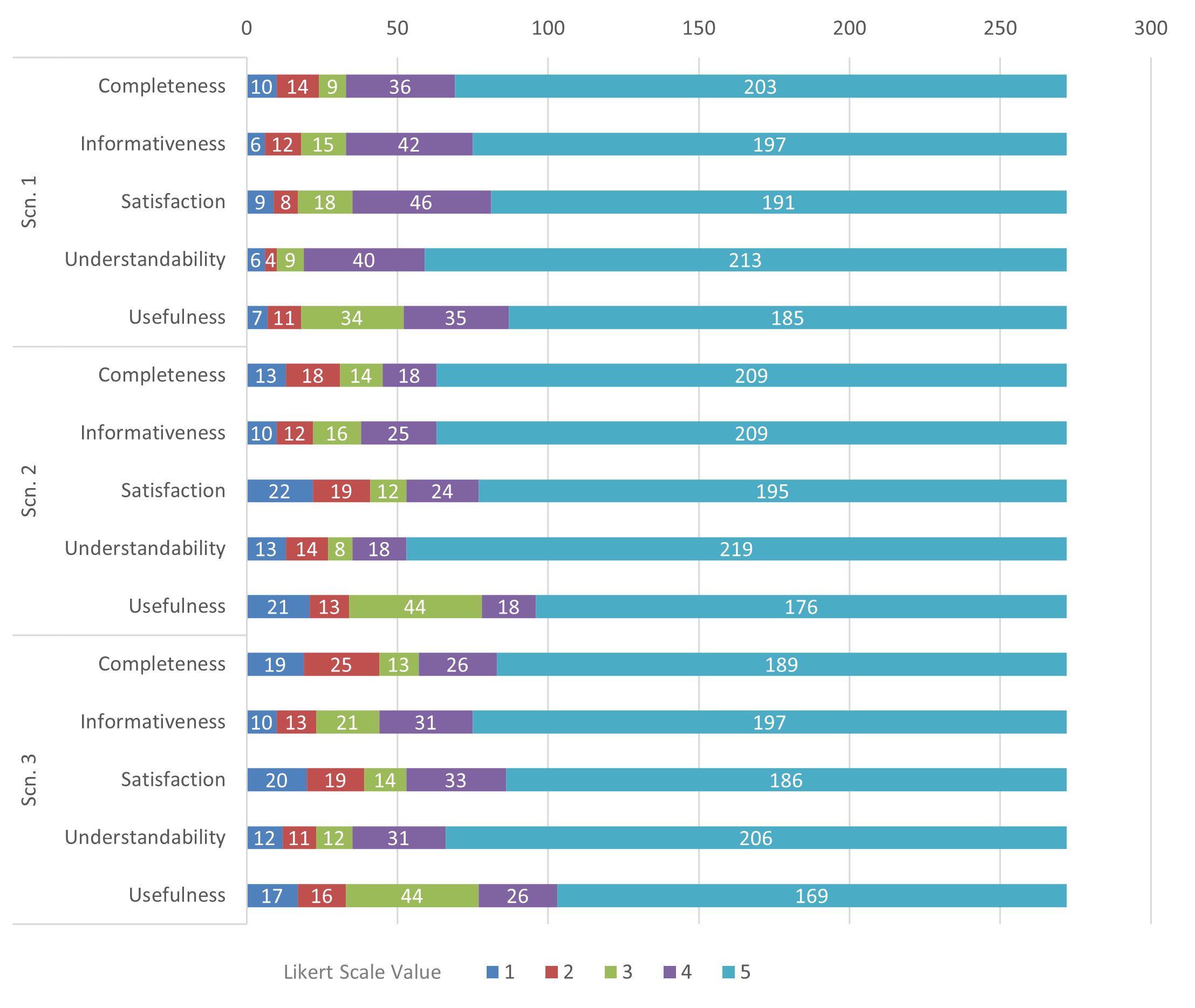}
	\caption{Distribution of Likert scale responses across scenarios and evaluation criteria.}
	\label{fig:humanEval_DataDist}
\end{figure*}

Figure \ref{fig:humanEval_DataDist} depicts the distribution of Likert scale responses for the five evaluation criteria across the three scenarios, showing the frequency of each response value. The highest ratings predominate across all scenarios and criteria, suggesting that participants generally found the explanations to be satisfactory. However, Scenarios 2 and 3 exhibit greater variability, with an increase in mid-range scores for certain criteria, such as Usefulness. These trends are particularly noticeable in broad questions, such as those in Category 1 regarding the navigation process overview, or when the model's answers do not align with the ground truth. In such cases, participant opinions diverge significantly, even in the presence of factual inaccuracies, indicating a potential bias in how utility is perceived when discrepancies with the ground truth arise.

\subsubsection{Reliability Analysis}
The reliability analysis resulted in an overall Cronbach's Alpha value of 0.9792, indicating good internal consistency and confirming that the evaluation criteria reliably capture the intended dimensions of the robot’s explanations. 

Additionally, we calculated Cronbach's Alpha if an item is deleted for each criterion. This metric helps to assess the impact of removing an individual item on the scale's overall reliability. As summarized in Table \ref{table:CronbachHumanEval}, the "Alpha if Deleted" values are consistently close to the overall Cronbach's Alpha. Specifically, these values range from 0.9699 (Usefulness) to 0.9798 (Informativeness), with all values being slightly lower than the overall result. This suggests that removing any single item would have only a minimal impact on the reliability of the scale, indicating that each criterion contributes positively to the measurement of overall quality.

Further, we examined item-total correlations, which quantify the strength of the relationship between each criterion and the total score. These correlations range from 0.9286 for the Informativeness criteria to 0.9660 for Usefulness. The high item-total correlations support the relevance of each item in assessing the key dimensions of the robot’s explanations.

\begin{table}[ht!]
\centering
\caption{Cronbach's Alpha and item-wise "Alpha if Deleted" in human evaluation.}
\begin{tabular}{lcc}
\toprule
 Criteria & Cronbach's Alpha if Deleted & Item-Total Correlation  \\
\midrule
Completeness & 0.9724  & 0.9492 \\
Informativeness & 0.9798  & 0.9286 \\
Satisfaction & 0.9752 & 0.9567 \\
Understandability & 0.9724 & 0.9502 \\
Usefulness & 0.9699 & 0.9660 \\
\bottomrule
\end{tabular}
\label{table:CronbachHumanEval}
\end{table}

Finally, we assessed inter-item correlations, which provide insight into how strongly the criteria correlate with one another. The inter-item correlation matrix, presented in Table \ref{table:CronbachHumanEvalInter}, shows values ranging from 0.878 (between Informativeness and Satisfaction) to 0.981 (between Satisfaction and Usefulness). These high correlations indicate that the criteria are closely related but not redundant, as each captures a unique and complementary aspect of the evaluation process.

\begin{table*}[ht!]
\centering
\caption{Inter-Item correlation matrix in human evaluation.}
\begin{tabular}{lccccc}
\toprule
 & Completeness & Informativeness & Satisfaction & Understandability & Usefulness  \\
\midrule
Completeness &  1.000 & 0.976 & 0.905 & 0.911 & 0.911  \\
Informativeness &  0.976 & 1.000 & 0.878 & 0.887 & 0.881\\
Satisfaction & 0.905 & 0.878 & 1.000 & 0.936 & 0.981\\
Understandability & 0.911 & 0.887 & 0.936 & 1.000 & 0.947 \\
Usefulness & 0.911 & 0.881 & 0.981 & 0.947 & 1.000 \\
\bottomrule
\end{tabular}
\label{table:CronbachHumanEvalInter}
\end{table*}

\subsubsection{Correlation Analysis}
Building on the findings of the reliability analysis, we further explored the relationships among the variables in our dataset using Pearson correlation matrices. While Cronbach's Alpha results demonstrated the internal consistency of the evaluation criteria, the correlation analysis extends it by examining how these criteria relate to each other across different scenarios and explanation categories.

\begin{figure}
	\centering
	\includegraphics[height=6.7cm]{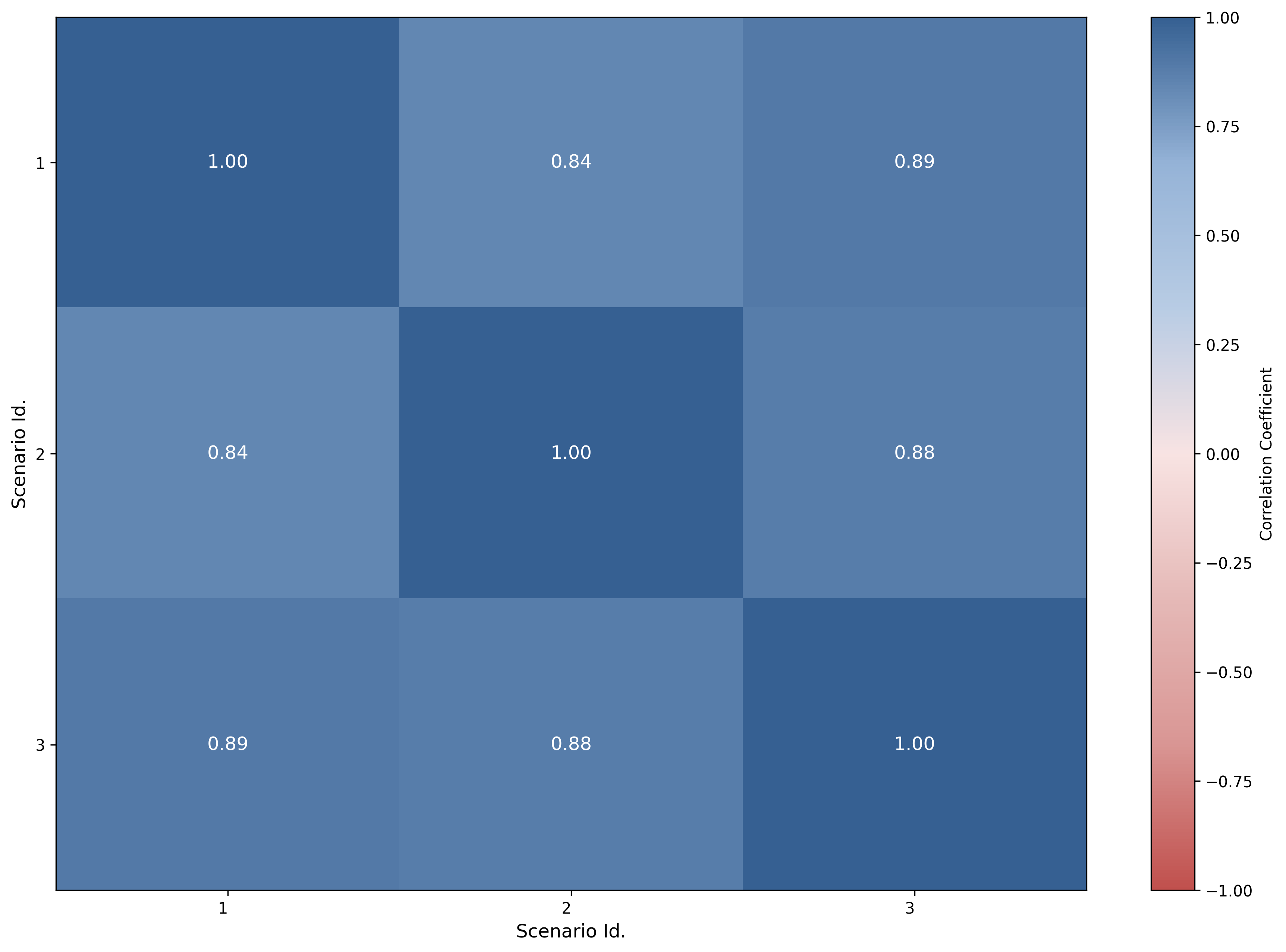}
	\caption{Correlation matrix of responses by scenario.}
	\label{fig:correlationMatrixScenario}
\end{figure}

Figure \ref{fig:correlationMatrixScenario} presents the Pearson correlation matrix for participants' responses across the three scenarios. This reveals strong associations, indicating a high level of consistency in the participants' assessments. This consistency suggests that the explanations provided were clear and easily comparable, allowing participants to reliably evaluate them regardless of the specific context of the robot's task.

\begin{figure}
	\centering
	\includegraphics[height=6.7cm]{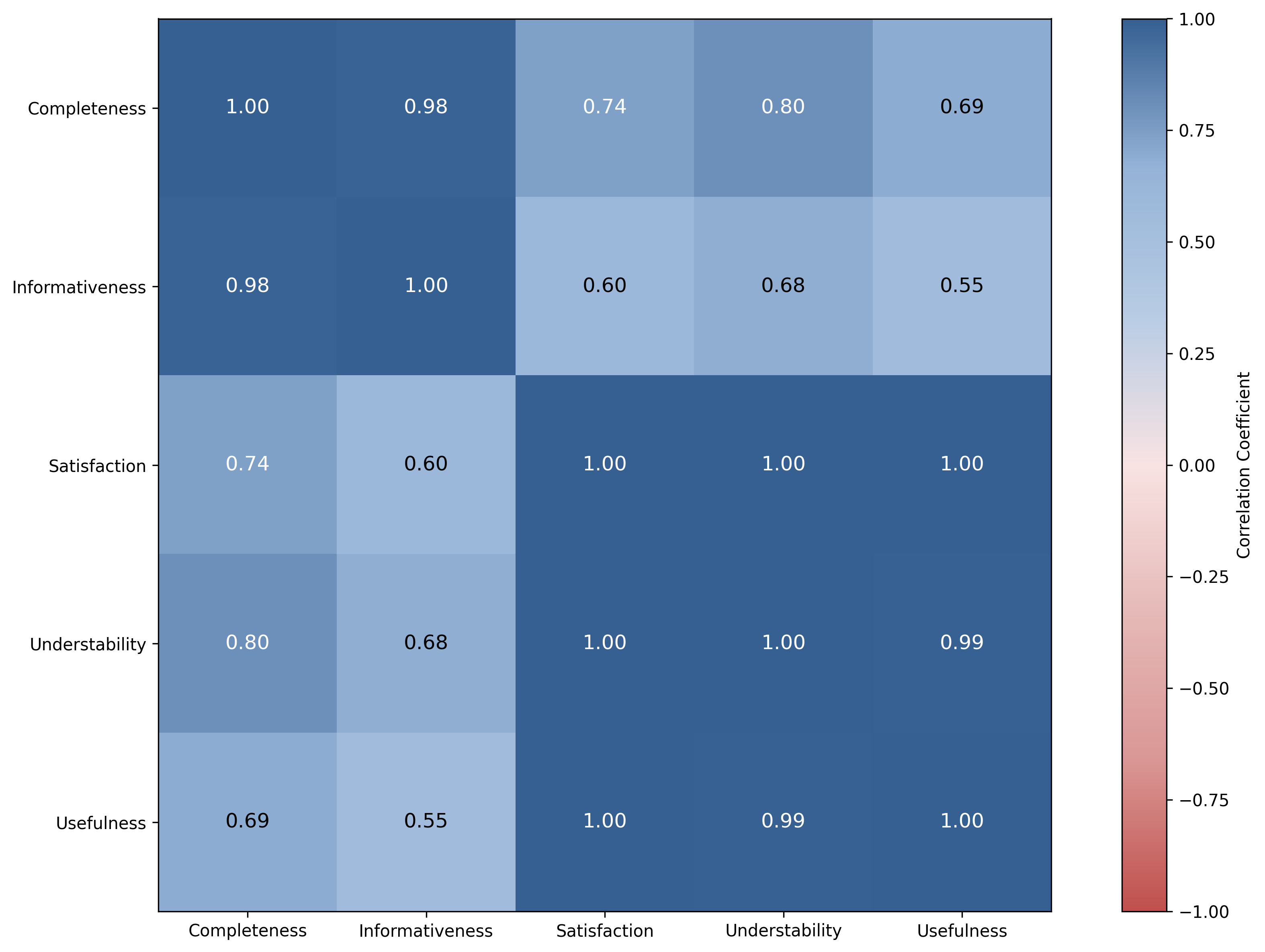}
	\caption{Correlation matrix of responses by criteria.}
	\label{fig:correlationMatrixCriteria}
\end{figure}

Figure \ref{fig:correlationMatrixCriteria} depicts the relationships among the evaluation criteria. Completeness and Informativeness exhibit a strong positive correlation, as do Understandability and Usefulness. This indicates that participants are more likely to perceive explanations as useful when they are clear. In contrast, the correlation between Informativeness and Usefulness is weaker (0.55), highlighting a divergence in participants' evaluations. This divergence may stem from the biases previously discussed, where participants find it challenging to judge the usefulness of an explanation that is not fully accurate.


\subsubsection{Inferential Statistics}
Statistical analysis often requires an understanding of the underlying data distribution to ensure that appropriate methods are applied. As parametric tests assume normality, violations of this assumption can compromise the validity of the results. To assess normality, we applied the Shapiro-Wilk test, a robust method for small to moderate-sized samples, across various groupings, including scenarios, categories, questions, and criteria.

The test results consistently revealed significant deviations from normality, with \(p\)-values well below 0.05, leading to rejection of the null hypothesis of normality for all groups. These findings confirm that the dataset does not satisfy the assumptions required for parametric tests, such as ANOVA or t-tests. As a result, non-parametric methods were adopted in subsequent analyses to ensure the robustness and validity of statistical inferences.

The Friedman test was used because of its ability to assess differences in related groups when the data does not meet the assumptions of normality required for the parametric tests. The results of the Friedman test across the three scenarios reveal significant differences in participant ratings for several explanations, indicating varying levels of agreement across conditions. In Scenario 1, significant differences were observed for the explanations of Questions 1 and 2, suggesting diverse responses. In contrast, most of the other questions in this scenario showed no significant differences (\(p \geq 0.05\)). 

Explanations 1 and 2 in all three scenarios addressed broad questions, whose answers diverged from the ground truth, as discussed in the GPT-4 evaluation in \ref{correctnessLLMEval}. Broad questions often result in responses that prioritize varied information, such as BTs, locations, or timestamps, making it difficult to align the answers with the ground truth. This fact justifies the significant variability in the participant ratings for these explanations.

In Scenario 2, significant differences were observed for Explanations 1, 7, 8, and 15, with \(p\)-values below the threshold of 0.05, while the other questions did not show significant variations. Specifically, the explanations for Questions 7 and 8 were misaligned with the ground truth, leading to significant variability in the participant assessments. In these cases, the participants' ratings lacked a clear trend across the evaluated criteria. A similar pattern was observed in Explanation 15 of Scenario 2, where the model provided irrelevant information that did not directly address the user's inquiry. This lack of relevance and alignment likely contributed to the observed variability in the participant ratings.

Finally, Scenario 3 showed significant differences for Explanations 1, 2, 7, and 16, indicating that the conditions for these explanations led to notable variations in the participant ratings. Explanation 7, does not address the answer to the corresponding question, including no relevant information. Explanation 16 in Scenario 3 caused variability in participants' answers due to its incomplete nature, highlighting the importance of content clarity in generating consistent and meaningful explanations. These findings suggest that while most explanations remain consistent, certain ones resonate differently depending on the scenario, nature of the question, and LLMs indeterminism.

\begin{figure*}
	\centering
	\includegraphics[height=8.5cm]{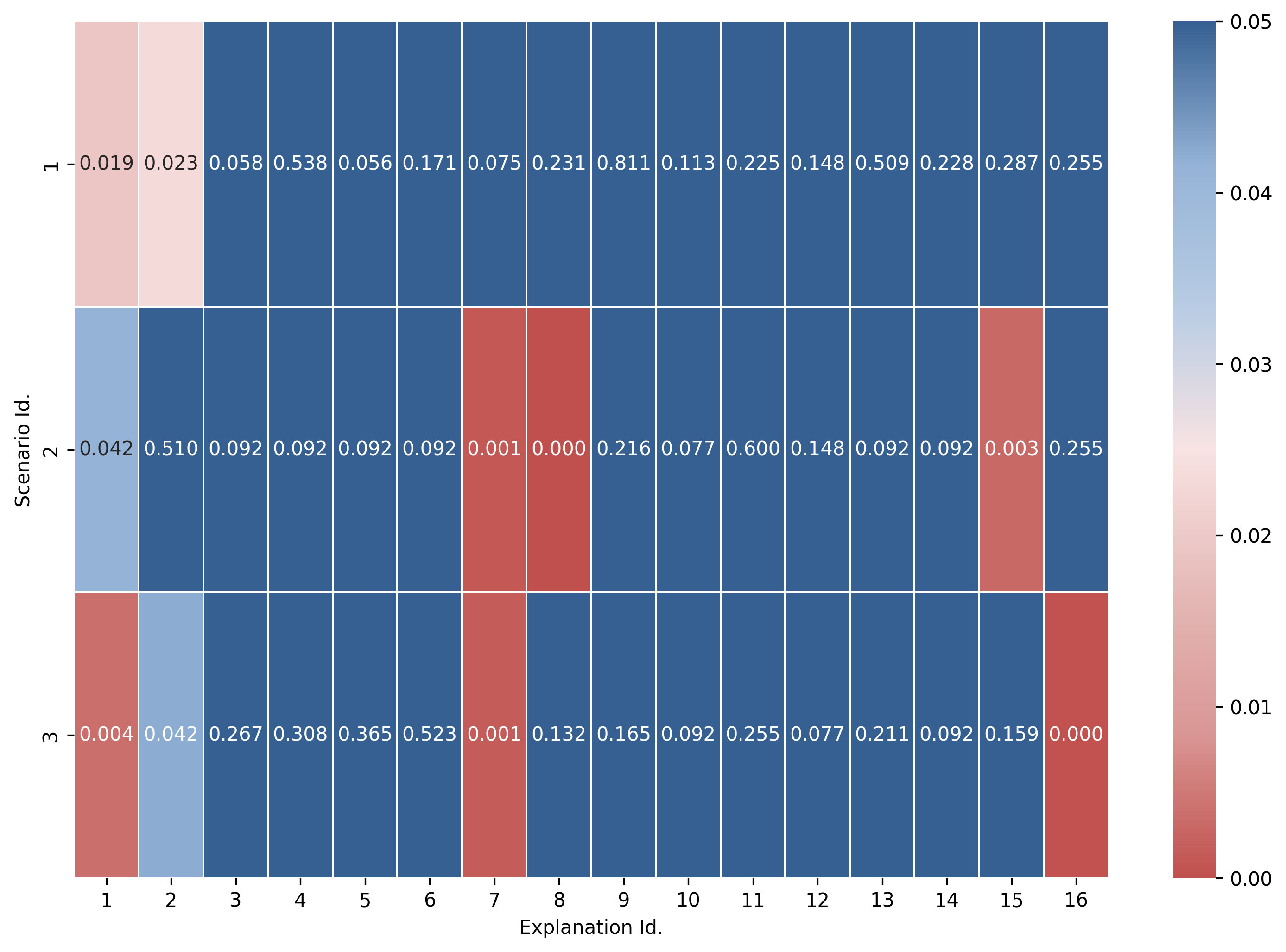}
	\caption{Friedman test results.}
	\label{fig:friedmanHetmap}
\end{figure*}

A post-hoc Nemenyi test was conducted for explanations that showed significant differences in participant ratings based on the Friedman test results. This allowed us to identify specific participant pairs with notable divergences in their responses across the three scenarios. The Nemenyi test revealed that most participants were consistent in their evaluations however, certain pairs exhibited significant variability. For example, in Scenario 1, the evaluation for Explanations 1 and 2 highlighted differences involving Participant 8 and several others. In Scenario 2, significant differences emerged for Explanation 1 between Participants 3, 6, 7, and 13. Similarly, in Scenario 3, variability was observed for Explanations 1 and 16, where the selected participants differed significantly from the others. These discrepancies are likely due to natural human biases, differences in interpretation, and subjective judgments inherent in human responses. Overall, while the explanations were generally clear and consistent for most participants, the observed variability underscores the role of individual perspectives and biases in shaping evaluations, reflecting the inherent complexity of human judgment.

\subsection{Similarity Between Human and LLM Evaluations}
As described in \ref{ExplanainabilityEvaluation}, previous studies have demonstrated that LLMs can effectively approximate human judgment in various evaluation tasks, including those involving subjective interpretation. In this study, we conducted a focused validation of LLM-generated evaluations by comparing a subset of 48 explanations from one of the six evaluation runs performed by the LLM against human assessments. This subset included 16 explanations per scenario, allowing us to validate the LLM judgment within a practical scope.

\begin{figure}[ht]
    \centering
    \subfigure[Human evaluation.]{%
        \includegraphics[height=6cm]{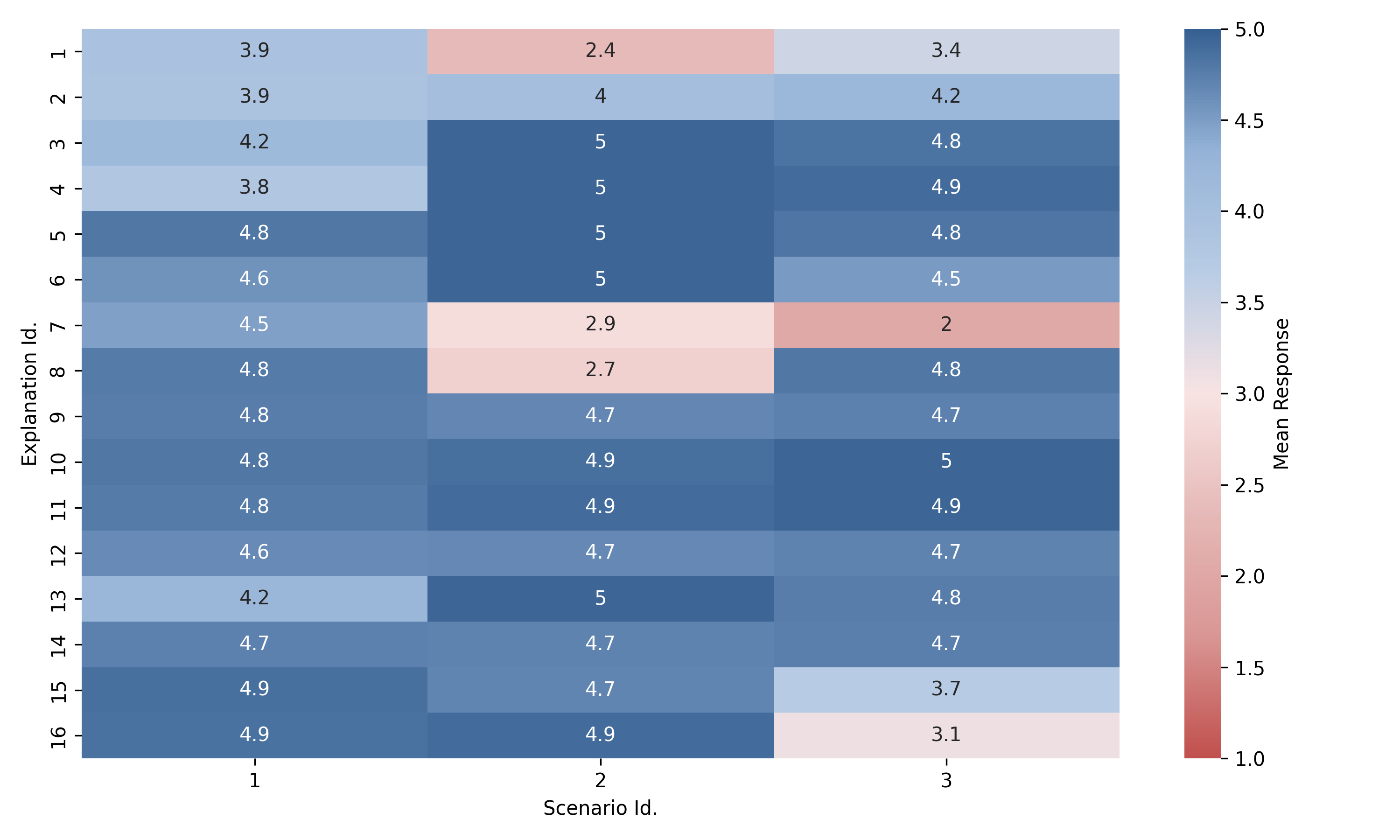}
        \label{fig:heatmapHumanEval}
    }\hfill
    \subfigure[LLM-as-Judge evaluation.]{%
        \includegraphics[height=6cm]{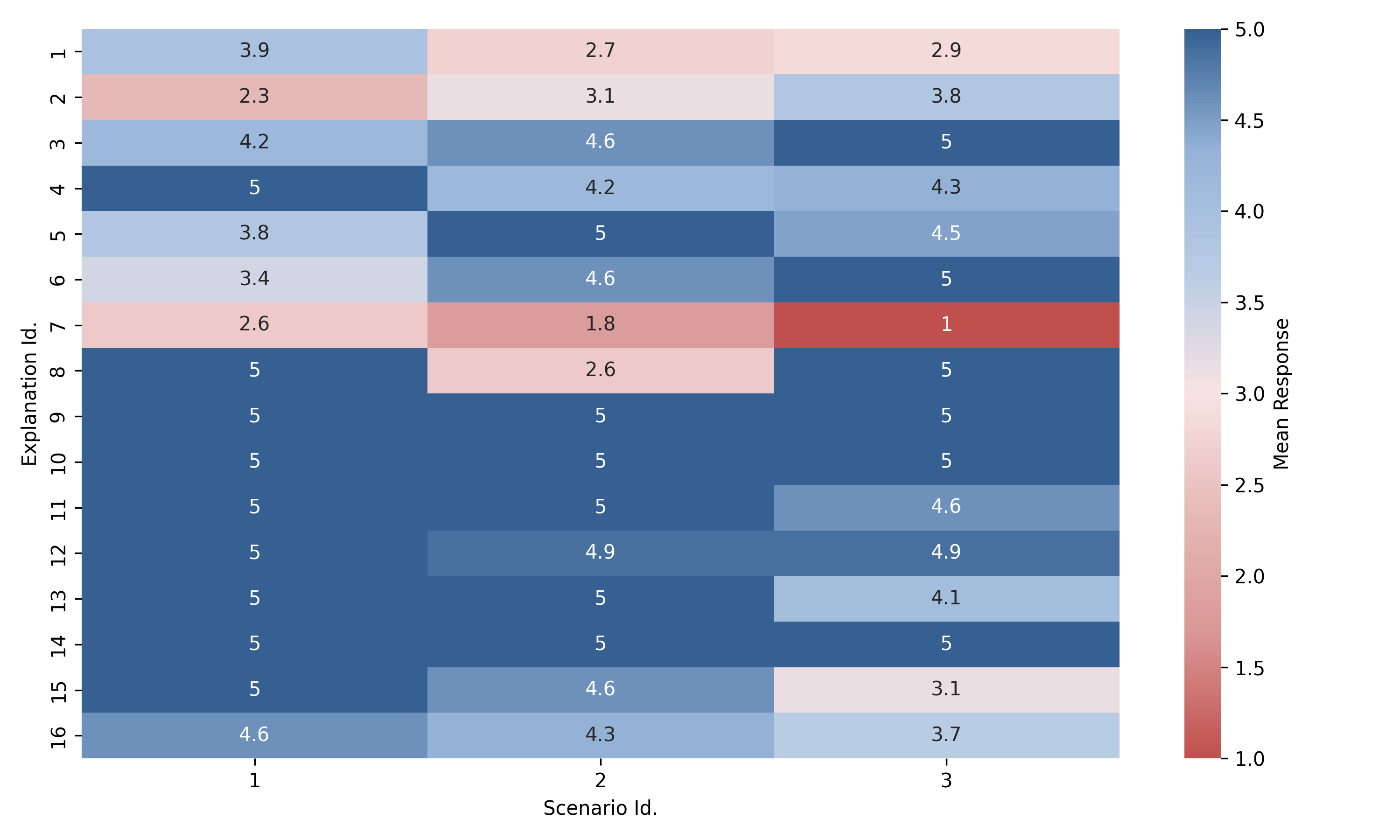}
        \label{fig:heatmapLLMEval}
    }
    \caption{Heatmaps of mean responses across scenarios and explanations. (a) Human evaluators. (b) LLM-as-Judge evaluation.}
    \label{fig:heatmapsHumanVsLLM}
\end{figure}

Figures \ref{fig:heatmapHumanEval} and \ref{fig:heatmapLLMEval} depict the mean responses from the 17 human participants and GPT-4, respectively, acting as a judge in this subset. The comparison highlights notable similarities in scoring patterns across scenarios and explanations, suggesting a high degree of alignment between human and LLM evaluations. Both evaluations consistently identify high-quality explanations with scores clustered around 4.5–5, indicating a shared agreement on the quality of certain explanations. Similarly, lower-rated explanations identified by humans are also flagged by the LLM, revealing mutual recognition of weaker content.

However, although the overall trends are similar, there are some differences in the exact ratings and outlier values between the two evaluations. These differences could be attributed to nuances in how the LLM interprets the explanations compared to human evaluators, who might consider subtle contextual or subjective factors that an LLM might miss. Despite these slight variations, the visual correspondence between the human and LLM heatmaps indicates that the LLM provides a reliable approximation of human judgment, making it a promising tool for scaling evaluation processes without sacrificing much in terms of accuracy. This similarity could be valuable in demonstrating that LLMs are suitable alternatives for tasks like explanation evaluation, especially in large-scale studies where human evaluation may be impractical.

\section{Conclusions}\label{sec:conclusions}


This work presents an accountability and explainability architecture for Robot Operating System (ROS)-based mobile robots. The proposed system comprises a black box component that ensures the verifiability and integrity of logged messages using blockchain technology, and an explainability component based on Retrieval-Augmented Generation (RAG) and Large Language Models (LLMs). This private question-answering system leverages the data recorded by the accountability component to provide context for generating natural language explanations. The results demonstrate the feasibility of producing accurate, coherent, concise, and helpful explanations from raw Rosbag data, thereby supporting user understanding and enhancing human-robot interactions.

Both components are integrated through a third main module in charge of filtering non-relevant information and transforming technical content into more accessible messages. This functionality reduces the complexity for developers by eliminating the need for direct interaction with the logging Application Programming Interface, thereby easing the development process and improving the adaptability and scalability of the system across diverse navigation tasks.

Our evaluation confirms that the system guarantees data immutability without compromising performance, supporting post-hoc analysis and failure investigation. Specifically, hashing one of every 100 messages resulted in less than 5\% message loss, even for high-frequency topics exceeding 300 Hz, with minimal resource consumption. These results validate the efficiency and practicality of accountability mechanisms.

The system also provides effective explanations across key aspects of autonomous navigation, such as goal completion, obstacle detection, trajectory re-planning, and task cancellation or abortion. The generated explanations achieved high scores for accuracy, coherence, and user satisfaction. LLM-as-a-judge evaluations reported correctness rates at or above 75\% across all scenarios, while human evaluations showed strong agreement, with a Cronbach’s Alpha of 0.9792, indicating high internal consistency. The alignment between human and LLM judgments suggests that LLMs can serve as reliable and scalable alternatives to human evaluation for assessing explanation quality.

These findings highlight the potential of the proposed architecture to support the broader deployment of autonomous agents in safety-critical scenarios. By enabling reliable post-hoc inspection and producing accessible, high-quality explanations, our system bridges the gap between complex robotic behavior and human interpretation. This capability is particularly relevant in contexts that require transparency, positioning our approach as a practical foundation for responsible and explainable robots.

While our solution has shown promising results in navigation functionalities across various scenarios, future work should extend its application to other domains by tailoring the recording and processing components to specific topics relevant to each context. Furthermore, the consistency and robustness of the responses generated by our explainability approach could benefit from refinement across different scenarios, especially when categorizing questions under specific subjects. To this end, future efforts should explore the impact of In-Context Learning techniques on prompt formulation, along with fine-tuning the model to better align with the unique demands of autonomous agent tasks.

Additionally, our RAG approach could be further optimized to retrieve more relevant information from the available context. This improvement may involve structuring the vector database into separate collections tailored to specific information types or adopting a hybrid retrieval strategy that combines Knowledge Graphs with vector-based indexing. Incorporating real-time explanation generation into the architecture could also enhance the responsiveness and adaptability of the system. Together, these enhancements have the potential to significantly improve the accuracy, relevance, and consistency of the generated explanations across diverse scenarios, thereby strengthening the overall effectiveness of the proposed architecture.

\appendices

\section{Supplementary Materials}

\subsection{Prompt Template}

\begin{figure}
	\centering
	\includegraphics[height=4cm]{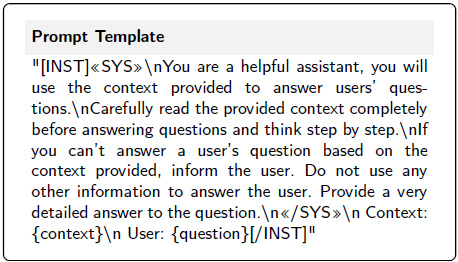}
	\caption{Prompt template used to generate explanations with Llama 2 LLM.}
	\label{PromptTemplate}
\end{figure}

\subsection{Explanations Obtained}
The 288 questions and answers obtained through the explanation generation and evaluation process are available in the aforementioned mentioned Github repository \footnote{https://github.com/laurafbec/immutable\_explainable\_BBR/blob/main/ \\ gpt4\_eval\_results/LLMs\_QA\_Results.xlsx}. A Python notebook has been developed to interact with the content of the previous sheet. This notebook is available at the following link \footnote{https://github.com/laurafbec/immutable\_explainable\_BBR/blob/main/\\gpt4\_eval\_results/GPT4 Evaluation Results.ipynb}.

\section*{Acknowledgment}

This publication is part of the TESCAC project, financed ``by European Union NextGeneration-EU, the Recovery Plan, Transformation and Resilience, through INCIBE''.  In addition, this work has been partially funded by the project EDMAR, PID2021-126592OB-C21, funded by 
MCIN/AEI/10.13039/501100011033 and by ERDF A way of making Europe.

\bibliography{refs}

\begin{IEEEbiography}[{\includegraphics[width=1in,height=1.25in,clip,keepaspectratio]{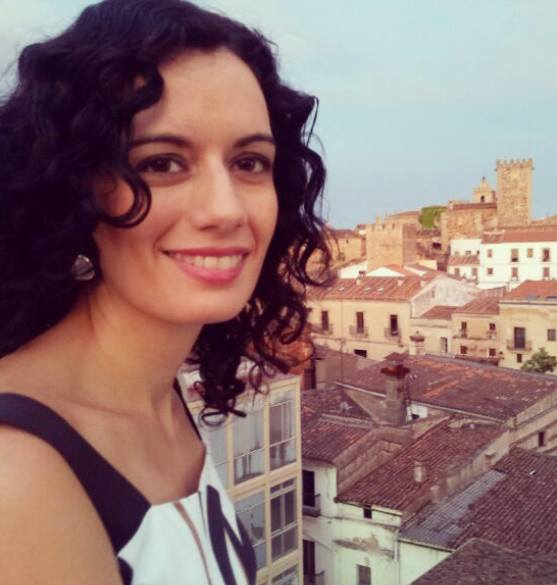}}]{Laura Fernández-Becerra} graduated in Computer Science (2005) at University of León (Spain). She also obtained a master of science (2012) in ICT Management at the University of Extremadura (Spain) and a master of science (2021) in Cybersecurity at the University of León. Since 2008 she has been working as a Systems Engineer at the Regional Government of Cáceres and is currently pursuing a Ph.D. in Computer Science at the University of León. Her research interests focus on cybersecurity, accountability and explainability in autonomous agents, and AI.
\end{IEEEbiography}

\begin{IEEEbiography}[{\includegraphics[width=1in,height=1.25in,clip,keepaspectratio]{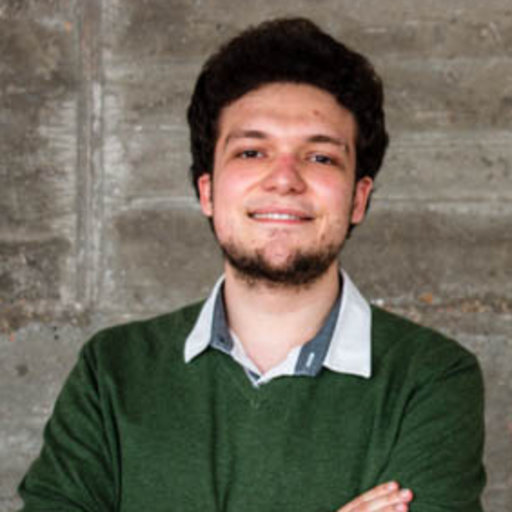}}]{Miguel Á. González-Santamarta} received his PhD degree in robotics in 2024 from the School of Industrial Engineering and Information Technology at the University of León (Spain). Currently, he is working as a researcher in the M Robotics Group at the University of León (Spain).  He has been a Research Associate in the Robotics Group at the Universidad de León since 2018. His research interests include cognitive robotics, cognitive architectures, explainability in autonomous robots and space robotics.
\end{IEEEbiography}

\begin{IEEEbiography}[{\includegraphics[width=1in,height=1.25in,clip,keepaspectratio]{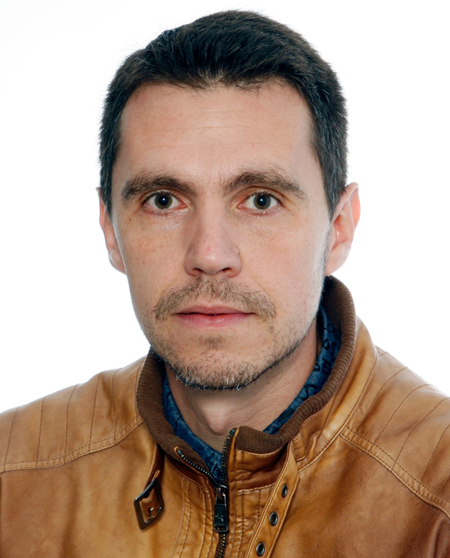}}]{Ángel Manuel Guerrero-Higueras} got his degree (2007) and his master of science (2010) in computer science at Rey Juan Carlos University (Madrid, Spain). Besides, he got his Ph.D. at the University of León in 2017.  In academia, he worked as a research assistant in the Atmospheric Physics Group (2011–2013) and in the Research Institute of Applied Science to CyberSecurity (2016–2018), both depending on the University of León. He currently stands as an Assistant Professor at the University of León. His main research interests include cybersecurity, robotics, and AI.
\end{IEEEbiography}

\begin{IEEEbiography}[{\includegraphics[width=1in,height=1.25in,clip,keepaspectratio]{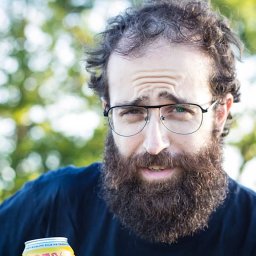}}]{Francisco J. Rodríguez-Lera} received his Ph.D. degree in intelligent systems for engineering in 2015 from the School of Industrial Engineering and Information Technology at University of León (Spain). Currently, he is working as a researcher in the Mobile Robotics Group at University of León (Spain) after two years working as a postdoctoral research associate in the AI Robolab, which belongs to the Computer Science and Communications Research Unit (CSC) at University of Luxembourg. His research interests focus on developing technologies for social robots that can interact and communicate with people in real human-robot interaction scenarios. Specifically, he works on cybersecurity in autonomous systems, solutions for robots in public spaces, designing motivational cognitive architectures for natural robot behaviors, software development and DevOps for research and business, and AI applications for context-awareness in human-robot interaction environments.
\end{IEEEbiography}

\begin{IEEEbiography}[{\includegraphics[width=1in,height=1.25in,clip,keepaspectratio]{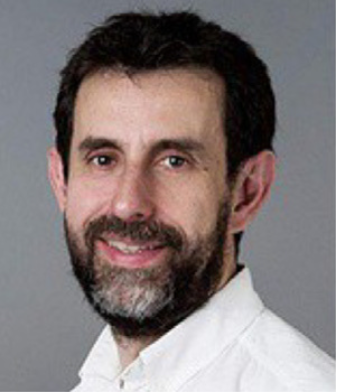}}]{Vicente Matellán-Olivera} got his PhD in Computer Science from the Technical University of Madrid in 1993. He was Assistant Professor at Universidad Carlos III de Madrid (Spain) (1993-1999). Associate Professor at Universidad Rey Juan Carlos (Spain) from 1999-2008. In 2008 he joined Universidad de León (León), where he still serves as Full Professor in the Mechanical, Computer and Aerospace Engineering Department. His main research interests have to do with robotics, artificial intelligence, and cybersecurity areas where he has made more than 250 contributions in journals, books, and conferences.
\end{IEEEbiography}



\EOD

\end{document}